\pdfoutput=1

\documentclass[11pt]{article}

\usepackage{amsmath,amsfonts,bm}

\def\1{\bm{1}}

\def\valpha{{\bm{\alpha}}}

\def\vomega{{\bm{\omega}}}

\def\vc{{\bm{c}}}

\def\vv{{\bm{v}}}

\def\vx{{\bm{x}}}
\def\vy{{\bm{y}}}

\def\mX{{\bm{X}}}

\DeclareMathAlphabet{\mathsfit}{\encodingdefault}{\sfdefault}{m}{sl}
\SetMathAlphabet{\mathsfit}{bold}{\encodingdefault}{\sfdefault}{bx}{n}

\def\gO{{\mathcal{O}}}

\def\gS{{\mathcal{S}}}

\def\sR{{\mathbb{R}}}

\newcommand{\softmax}{\mathrm{softmax}}

\DeclareMathOperator*{\argmin}{arg\,min}

\DeclareMathOperator{\RCS}{RCS}
\DeclareMathOperator{\ACS}{ACS}
\DeclareMathOperator{\DF}{DF}

\usepackage[final]{acl}

\usepackage{times}
\usepackage{latexsym}
\usepackage{subfigure}
\usepackage{graphicx}
\usepackage{booktabs}
\usepackage{xcolor}
\usepackage{enumitem}
\usepackage{caption}
\usepackage{subcaption}
\usepackage{comment}
\usepackage{algorithmic}
\usepackage{algorithm}
\usepackage{multirow}
\usepackage{url}

\usepackage[T1]{fontenc}

\usepackage[utf8]{inputenc}

\usepackage{microtype}

\usepackage{inconsolata}

\makeatletter
\newcommand{\clearsubcaptcounter}{\setcounter{sub\@captype}{0}}
\makeatother

\newcommand{\yh}[1]{\textcolor{blue}{#1}}

\newcommand{\astmark}[1][*]{\textsuperscript{#1}}

\AtBeginDocument{%
  \setlength\abovedisplayskip{0.6ex}
  \setlength\belowdisplayskip{0.6ex}}
\allowdisplaybreaks[4]

\title{CoBa: Convergence Balancer for Multitask Finetuning of\\Large Language Models}

\author{Zi Gong\thanks{Equal contribution.} \\ Ant Group \\ \texttt{gongzi.gz@antgroup.com} \And
        Hang Yu\astmark \\ Ant Group \\ \texttt{hyu.hugo@antgroup.com} \And
        Cong Liao \\ Ant Group \\ \texttt{liaocong.lc@antgroup.com} \AND
        Bingchang Liu \\ Ant Group \\ \texttt{bingchang.lbc@antgroup.com} \And
        Chaoyu Chen \\ Ant Group \\ \texttt{chris.ccy@antgroup.com} \And
        Jianguo Li\thanks{The corresponding author.} \\ Ant Group \\ \texttt{lijg.zero@antgroup.com}
        }

\begin{document}
\maketitle
\begin{abstract}

Multi-task learning (MTL) benefits the fine-tuning of large language models (LLMs) by providing a single model with improved performance and generalization ability across tasks, presenting a resource-efficient alternative to developing separate models for each task. 
Yet, existing MTL strategies for LLMs often fall short by either being computationally intensive or failing to ensure simultaneous task convergence. This paper presents CoBa, a new MTL approach designed to effectively manage task convergence balance with minimal computational overhead. Utilizing Relative Convergence Scores (RCS), Absolute Convergence Scores (ACS), and a Divergence Factor (DF), CoBa dynamically adjusts task weights during the training process, ensuring that the validation loss of all tasks progress towards convergence at an even pace while mitigating the issue of individual task divergence. The results of our experiments involving four disparate datasets underscore that this approach not only fosters equilibrium in task convergence but enhances the LLMs' performance by up to 13\% relative to the second-best baselines. Code is open-sourced at \url{https://github.com/codefuse-ai/MFTCoder}.

\end{abstract}

\section{Introduction}
\label{sec:introduction}

\begin{table}[t]
    \centering
    \caption{The time complexity of existing MTL approaches for each heterogeneous batch. In this context, `$F$' and `$B$' denote the time complexity of the forward and backward propagation respectively. `$K$' is the number of tasks. $\lvert\theta_{s}\rvert$ is the weights (usually the final layer of weights which are shared between tasks). The constants are referred to as `$a_{i}$', where $a_{4}<a_{1}<a_{2}<a_{3}$. Additionally, $^*$ means the loss weight is determined by the convergence trend of the validation rather than the training loss.}
    \vspace{-1ex}
    \resizebox{\linewidth}{!}{
    \begin{tabular}{lcccccc}
        \toprule
        \pmb{Method} & & & & & & \pmb{Time Complexity} \\
        \midrule
        Uniform & & & & & & $\gO(F+B)$ \\
        GradNorm & & & & & & $\gO(F+B+K\lvert\theta_{s}\rvert)$ \\
        GradNorm$^{*}$ & & & & & & $\gO(2F+2B+K\lvert\theta_{s}\rvert)$ \\
        LBTW & & & & & & $\gO(F+B+a_{1}K)$ \\
        LBTW$^{*}$ & & & & & & $\gO(2F+B+a_{1}K)$ \\
        FAMO & & & & & & $\gO(F+B+a_{2}K+a_{4}K^{2})$ \\
        FAMO$^{*}$ & & & & & & $\gO(2F+2B+a_{2}K)$ \\
        MetaWeighting & & & & & &  $\gO(KF+KB)$ \\
        CoBa & & & & & & $\gO(2F+B+a_{3}K)$ \\
        \bottomrule
    \end{tabular}}
    \label{tab:time-complexity}
    \vspace{-3ex}
\end{table}

In recent years, large language models (LLMs) have emerged as a focal point of research within both academia and industry, owing to their superior performance. These models are initially pretrained, designed to ensure they possess broad applicability across a variety of downstream tasks. This is followed by a finetuning stage, which meticulously adapts the models for specific tasks or scenarios. However, this phase requires individual, task-specific finetuning, leading to a complex deployment scenario in production environments. The need to deploy separate models for each task, combined with their considerable size and the associated resource consumption, presents a formidable challenge as the number of tasks grows.


Multi-task learning (MTL) presents a promising remedy to the above issue by enabling the simultaneous training of multiple tasks~\cite{crawshaw2020multi,vandenhende2021multi,zhang2023survey}. This approach leverages a single model to support a variety of tasks, thus significantly conserving resources. Moreover, MTL not only fosters performance improvements across related tasks but has the potential to generalize to unseen tasks. Reversely, the vast parameter space of LLMs facilitates this adaptability, allowing them to undertake multiple tasks simultaneously. This proficiency is exemplified by GPT-3.5/4 from OpenAI~\cite{achiam2023gpt}.

For an effective implementation of MTL in LLMs, two critical criteria must be met concurrently. First, \textbf{the approach should incur minimal extra computational costs} since the training of LLMs in itself is already highly resource-intensive. Second, \textbf{it is imperative to guarantee the simultaneous convergence of all tasks}, tactfully navigating to a shared optimal checkpoint.



Unfortunately, current approaches do not simultaneously meet the above two requirements. Traditional MTL methods, particularly those focusing on loss balancing and gradient manipulation~\cite{kendall2018multi,chen2018gradnorm,liu2019loss,mao2022metaweighting,liu2024famo}, have proven effective for smaller models and straightforward classification tasks. However, adapting these established techniques to LLMs presents significant challenges due to the high computational costs and the complexities involved in integrating them with parallel training frameworks. For example, GradNorm~\cite{chen2018gradnorm}, FAMO~\cite{liu2024famo}, and MetaWeighting~\cite{mao2022metaweighting} typically incur a high computational cost with regard to the number of tasks $K$, as shown in Table~\ref{tab:time-complexity}. Conversely, NLP models such as Muppet~\cite{aghajanyan2021muppet} and ExT5~\cite{aribandi2021ext5} employ a straightforward data mixing strategy from multiple tasks for application in LLMs. However, they fall short of addressing the persistent issue of uneven task convergence within MTL settings. This imbalance can result in a scenario where some tasks are still optimizing while others begin to worsen, negatively impacting the model's overall effectiveness.

In this paper, we introduce \textbf{CoBa} (COnvergence BAlancer), an innovative MTL approach designed for LLMs. This method aims to achieve balanced convergence across various tasks while maintaining ease of applicability in the training of LLMs. The core strategy involves dynamically varying each task's training loss weight based on its convergence trends in the validation dataset. Two essential criteria underpin this method are: 1) \textbf{when the validation losses of all tasks consistently decline, the method lowers weights for those converging faster (experiencing steeper drops in validation losses) and increases weights for those converging more gradually (exhibiting less steep slopes).} 2) \textbf{For any tasks showing divergence—a signal of possible overfitting—their associated weights are decreased. On the other hand, weights are boosted for tasks that are steadily converging.} To move forward to these objectives, we introduce the Relative Convergence Score (RCS) to address the first criterion and the Absolute Convergence Score (ACS) for the second. A Divergence Factor (DF) is then applied to ascertain which score prevails in influencing the final weight allocation. Note that RCS, ACS, and DF are all efficiently computed, leveraging the validation loss slopes through normalization and softmax functions, making them not only computationally effective but also easily compatible with parallel training architectures. To summarize, the main contributions of our study are: 

\begin{itemize}[leftmargin=*, topsep=1pt,itemsep=1pt,partopsep=1pt,parsep=1pt]
    \item We introduce CoBa, a novel strategy designed to achieve balanced convergence across various tasks. CoBa is straightforward in its application to LLMs, bridging the gap between advanced MTL requirements and practical usability.
    \item We propose two new metrics — the RCS and the ACS — along with a DF. The former two cater to the aforementioned two criteria respectively, while the DF determines which metric primarily affects the final weight distribution.
    \item We validate the efficacy and efficiency of CoBa through extensive experiments and show that CoBa not only maintains balanced convergence across tasks but also achieves up to a 13\% relative performance improvement in comparison with the second-best baselines.
\end{itemize}



\section{Related Work}
\label{sec:related_work}

All Multi-Task Learning (MTL) approaches~\cite{crawshaw2020multi,vandenhende2021multi,zhang2023survey} are designed to foster positive knowledge transfer across tasks through parameter sharing, while simultaneously minimizing any potential negative transfer, often referred to as task interference. Our discussion here primarily revolves around optimization techniques within MTL, given their relevance to LLMs. We categorize the existing work into two distinct groups: classical MTL methods and MTL methods tailored for LLMs.

\vspace{-1ex}
\paragraph{Classical Methods}
Traditional MTL strategies aimed at addressing task imbalance from an optimization standpoint fall into two categories: gradient manipulation and loss balance. Gradient manipulation techniques~\cite{chen2020just,liu2020towards,yu2020gradient,liu2021conflict} create a composite update vector at each optimization step by amalgamating task gradients. This approach ensures local improvements across all tasks but can be computationally intensive, particularly for models with numerous parameters. Conversely, loss balancing methods dynamically adjust each task's weight during training based on predefined factors such as task uncertainty~\cite{kendall2018multi}, task difficulty prioritization~\cite{guo2018dynamic}, and random loss weighting~\cite{lin2021closer}. These methods are computationally more efficient but do not guarantee simultaneous convergence of all tasks. To overcome this issue, advanced solutions aiming at convergence balance include DWA~\cite{liu2019end}, LBTW~\cite{liu2019loss}, GradNorm~\cite{chen2018gradnorm},  MetaWeighting~\cite{mao2022metaweighting}, and FAMO~\cite{liu2024famo}. The latter three adjust task weights based on gradients, with MetaWeighting additionally focusing on the validation instead of the training loss to enhance generalization performance. Unfortunately, gradient-based weight adjustment can be computationally demanding, as shown in Table~\ref{tab:time-complexity}. 

\vspace{-1ex}
\paragraph{Methods for LLMs}
MTL research specific to LLMs is still in its infancy, with a handful of notable contributions from models such as T5~\cite{raffel2020exploring}, Muppet~\cite{aghajanyan2021muppet}, ExT5~\cite{aribandi2021ext5}, and MFTcoder~\cite{liu2023mftcoder}. The initial three primarily aggregate data from various tasks without considering task equilibrium, often overlooking tasks with smaller datasets and favoring those with larger ones. MFTcoder advances this by calculating individual loss for each task, yet assigns equal weights across the board. MFTcoder acknowledges the inability of such approaches to ensure uniform validation loss convergence across tasks and suggests leveraging FAMO as a potential solution.

The proposed method, CoBa, embodies the strengths of both classical and LLM-specific MTL approaches, achieving convergence balance among tasks with minimal additional computational demands. Crucially, it focuses on validation loss, thereby promising to maintain or improve the generalization capabilities of the model.

\begin{figure*}[t]
    \centering
    \subfigure{
    \includegraphics[width=0.4\linewidth,origin=c]{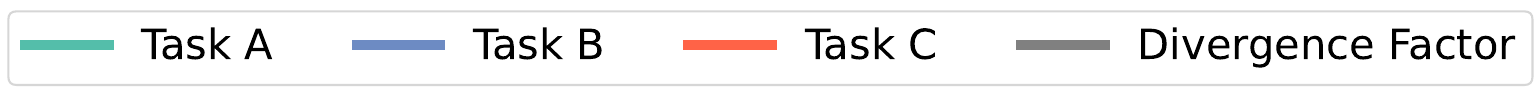}}\\
    \vspace{-2.5ex}
    \clearsubcaptcounter
    \subfigure[\scriptsize Loss ratios.]{
        \includegraphics[height=0.134\linewidth]{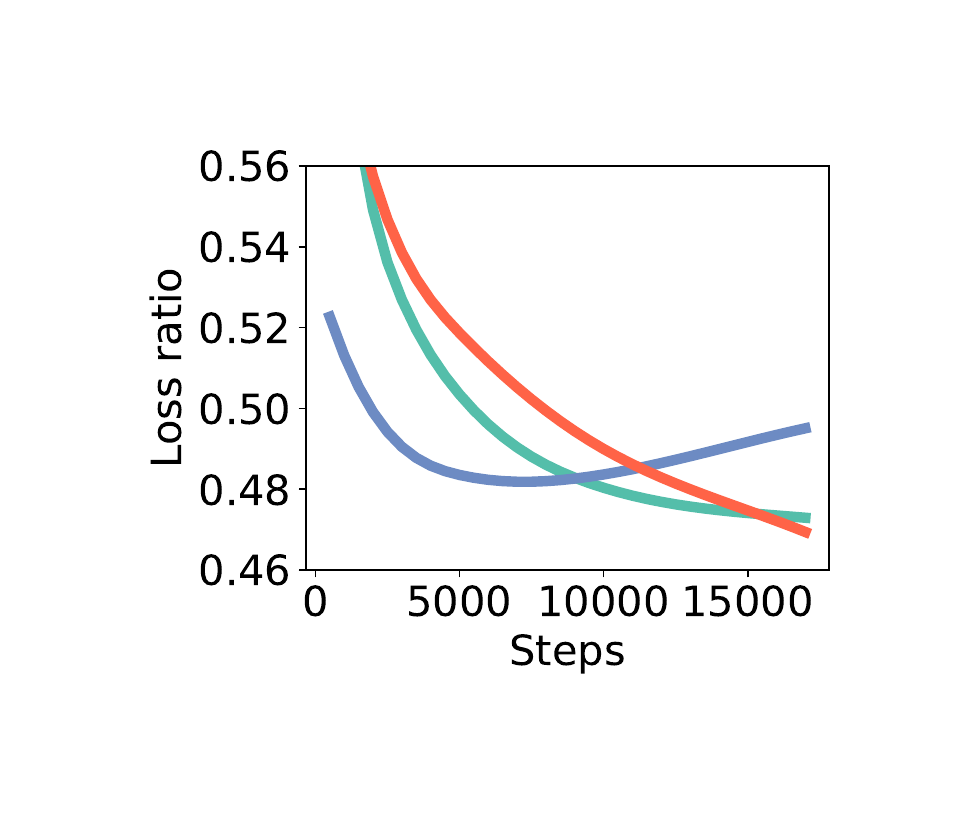}
        \label{fig1-a}
    }
    \hspace{-2ex}
    \subfigure[\scriptsize Convergence slopes.]{
        \includegraphics[height=0.13\linewidth]{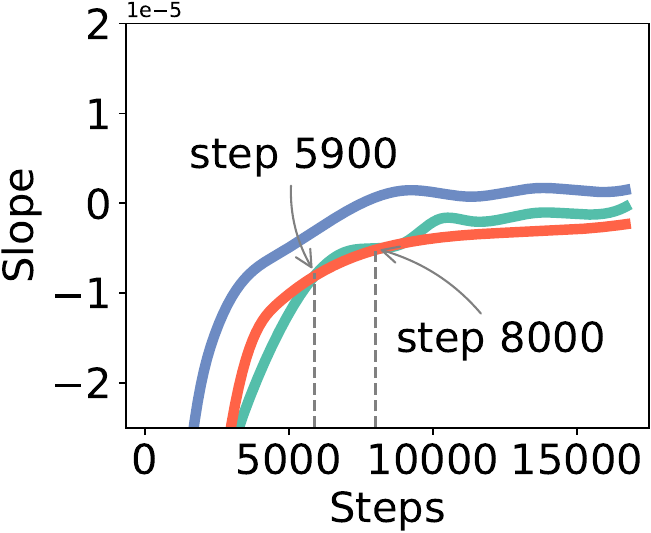}
        \label{fig1-b}
    }
    \hspace{-2ex}
    \subfigure[\scriptsize Relative scores.]{
        \includegraphics[height=0.13\linewidth]{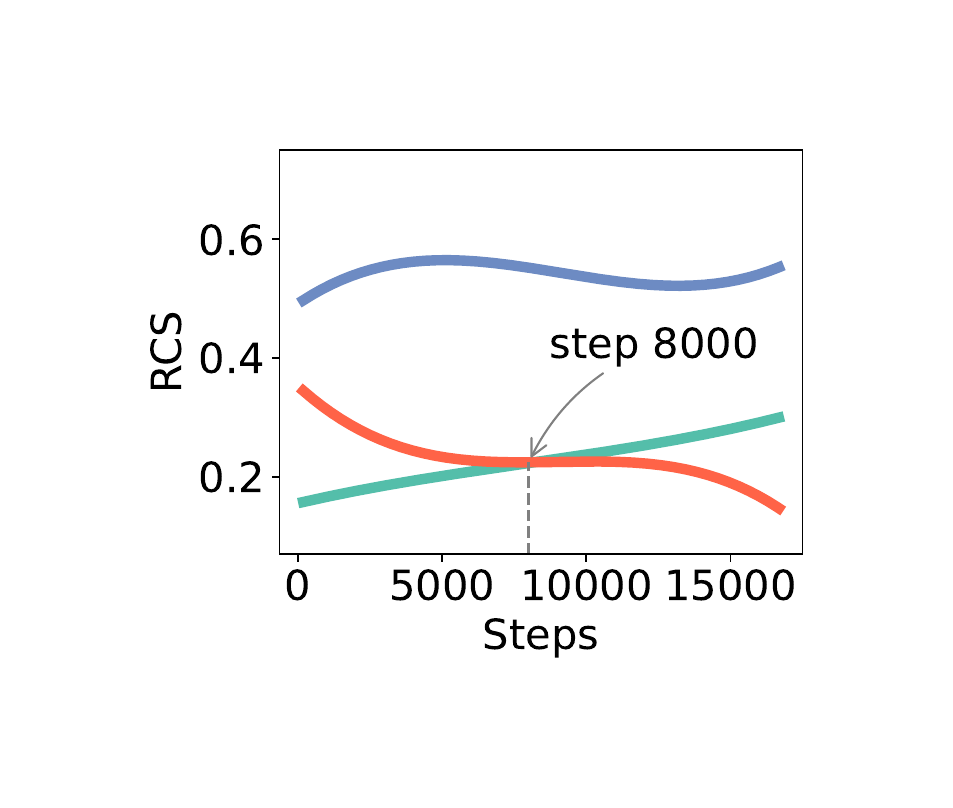}
        \label{fig1-c}
    }
    \hspace{-2ex}
    \subfigure[\scriptsize Absolute scores.]{
        \includegraphics[height=0.13\linewidth]{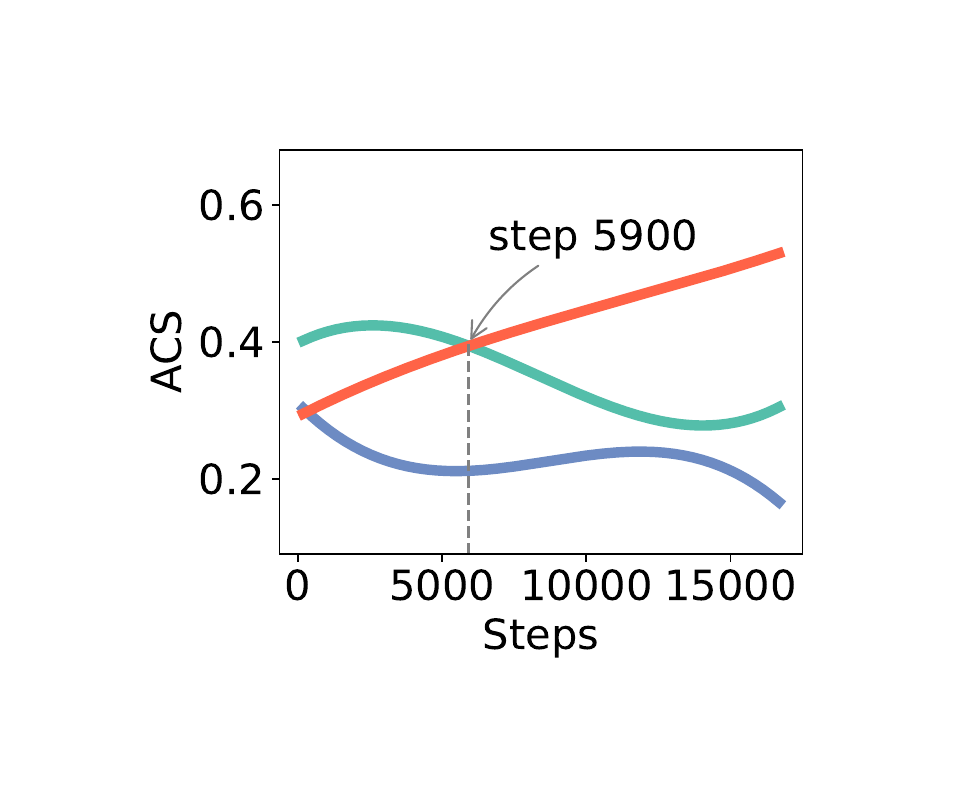}
        \label{fig1-d}
    }
    \hspace{-2ex}
    \subfigure[\scriptsize Divergence factor.]{
        \includegraphics[height=0.133\linewidth]{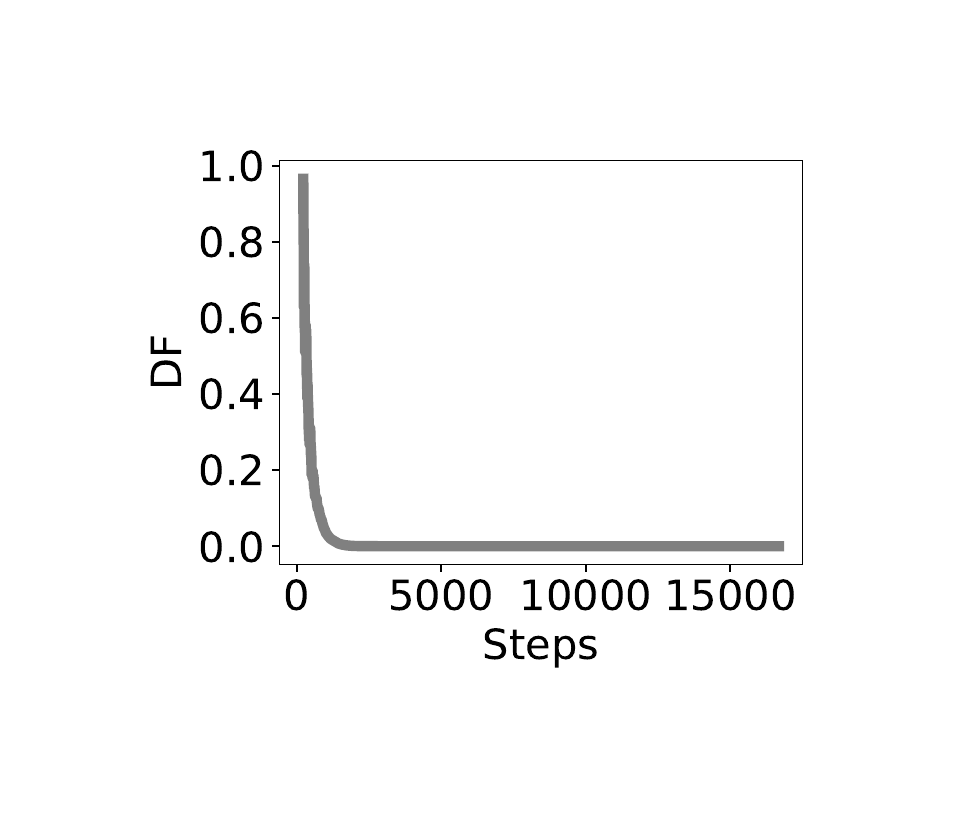}
        \label{fig1-e}
    }
    \hspace{-2ex}
    \subfigure[\scriptsize Weights.]{
        \includegraphics[height=0.13\linewidth]{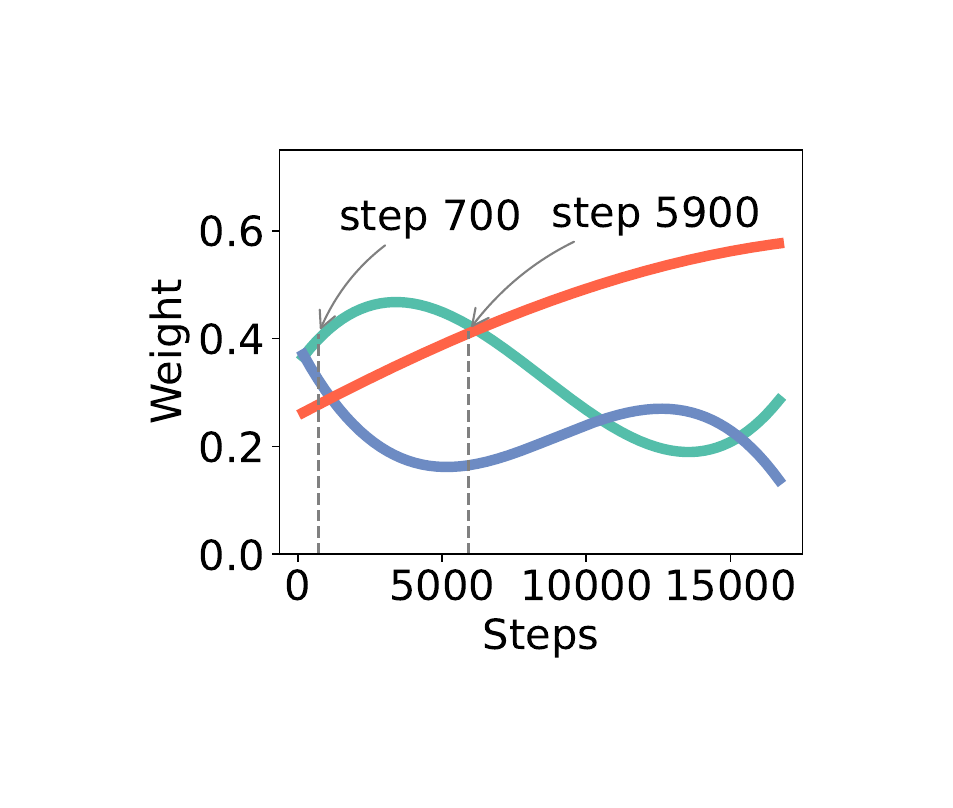}
        \label{fig1-f}
    }
    \vspace{-2ex}
    \caption{Demonstration of CoBa's task weight calculation process based on a real example.}
    \label{fig1}
    \vspace{-3ex}
\end{figure*}

\section{Convergence Balancer (CoBa)}
\label{sec:method}

Multi-task learning (MTL) is engineered to optimize a single model, parameterized by ${\theta} \in {\sR^m}$, enabling it to adeptly perform $K \geq 2$ tasks, potentially even in tandem. The loss function for task $i$ at the $t$-th iteration is denoted by ${\ell_i(\theta;t)}:{\sR^m}\rightarrow{\sR_{\geq 0}}$. This forms the foundation for the optimization challenge of MTL, expressed as:
\begin{equation}
    \mathop{\min}_{\theta \in \sR^m} \left\{\ell(\theta;t) := \mathop{\sum}_{i=1}^{K}{\omega_{i}(t) \ell_i(\theta;t)} \right\},
\label{eqa:1}
\end{equation}
where $\omega_{i}(t)$ is the loss weight for task $i$ at iteration $t$. Assigning $\omega_{i}(t) = 1/K$ ensures data balance, giving due attention to tasks irrespective of their sample sizes. However, this approach leads to varying convergence rates across tasks (e.g., see Figure~\ref{fig1-a}), complicating the identification of a checkpoint that is optimal for all tasks. Our goal is to adjust the weights $\omega_{i}(t)$ to harmonize these convergence rates. Furthermore, we prioritize generalization over mere training performance, and so the weights $\omega_{i}(t)$ is derived from the validation rather than training losses. To achieve these objectives, we adhere to two key criteria:

\textbf{\textit{c1}}.~When validation losses for all tasks are on a downward trend, tasks with faster convergence (sharper decline) receive reduced weights to avoid rapid overfitting. In contrast, tasks converging more slowly (gentler slopes) are assigned increased weights to encourage more learning.

\textbf{\textit{c2}}.~When any tasks begin to display signs of divergence (overfitting), their weights are decreased. Conversely, tasks that maintain a convergence trajectory are accorded higher weights.

These two criteria are quantified respectively through Relative Convergence Scores (RCS) and Absolute Convergence Scores (ACS). RCS is employed to assess the convergence pace relative among tasks, while ACS measures the current convergence rate against historical rates for each task individually. Note that these scores are derived from the slopes of validation losses—not the gradients—to minimize computational demands. We then integrate a divergence factor (DF), highlighting the overall convergence trajectory across tasks. This factor ensures RCS impacts weights predominantly when all tasks are converging, while ACS takes precedence as an arbitrary task commences diverging. In the sequel, we detail the computation of slopes, the formulation of RCS and ACS, and their amalgamation with the DF.

\subsection{Convergence Slope}

The convergence speed of different tasks can be intuitively measured by examining the slope of the validation loss curves. Figure~\ref{fig1-a} depicts this scenario, where Task B, marked by the green curve, demonstrates a quicker convergence compared to Task C, which is represented by the red curve, up until step 5900. This is reflected in the steeper slope observed for Task B, indicating a higher absolute value for its convergence slope than that of Task C as shown in Figure~\ref{fig1-b}.

To ensure a fair comparison of convergence speeds across tasks, we first normalize the validation losses. Specifically, we employ the validation loss ratio, $\Bar{\ell}_i^{val}(\theta;t)$, calculated as the current validation loss divided by the initial validation loss at step 0, that is, $\Bar{\ell}_i^{val}(\theta;t) = \ell_i^{val}(\theta;t)/\ell_i^{val}(\theta;0)$, where the $\ell_i^{val}(\theta;0)$ refers to the validation loss of the $i$-th task at step 0. 

Utilizing this normalized validation loss ratio, $\Bar{\ell}_i^{val}(\theta;t)$, we fit a linear model defined by $\alpha x + \beta$ across a selected range of iterations. The slope $\alpha$ from this linear fit provides us with an estimate of the convergence slope for that period. More specifically, at iteration $t$, we construct the observations vector $\vx_{i}(t) = [t,1]^\top$ and accordingly compile the observation matrix $\mX_{i}(N;t) = [\vx_{i}(s_0),...,\vx_{i}(t)]^\top$, matched with the corresponding validation loss ratios $\vy_{i}(N;t) = [\Bar{\ell}_i^{val}(\theta;s_0),...,\Bar{\ell}_i^{val}(\theta;t)]^\top$, where $\vx^\top$ denotes the transpose of $\vx$, $N$ refers to the length of history window, and $s_0 = \mathop{\max}(0, t-N+1)$. We aim to obtain the coefficient vector $\vc_i(N;t) = [\alpha_{i}(N;t),\beta_{i}(N;t)]^\top$, which minimizes the MSE between the projected values $\mX_{i}(N;t)\vc_{i}(N;t)$ and the actual values $\vy_{i}(N;t)$:
\begin{equation} 
\vc_{i} = \argmin_{\vc_{i}}\frac{1}{2}(\mX_{i}\vc_{i}-\vy_{i})^{\top}(\mX_{i}\vc_{i}-\vy_{i}). 
\label{eqa:6} 
\end{equation}
The vector $\vc_i(N;t)$ has a closed-form solution as:
\begin{equation} 
\vc_{i} = (\mX_{i}\mX_{i}^{\top})^{-1}\mX_{i}\vy_{i}^{\top}. \label{eqa:7} 
\end{equation}
Note that the solution in Eq.~\eqref{eqa:7} is only applicable for $t \geq 1$. Thus, we set $\alpha_{i}(N;t)=0$ when $t<1$.

Furthermore, to address the potential inaccuracy of initial convergence slopes, our methodology incorporates a warm-up mechanism, parameterized by $W$, which defines the number of steps before the weight update process begins. During this warm-up period, task weights are uniformly set to $1/K$, ensuring a balanced starting point. Once the warm-up period is completed, the weights are updated based on the convergence slopes observed within a sliding window of $N$ steps. We recommend setting $N$ to $2M$ and $W$ to $M$, where $M$ is the number of batches in the validation set. In each iteration, only a single mini-batch from the validation set is used for calculating the task-specific loss weight $w_i(t)$.

\subsection{Relative Convergence Scores (RCS)}
As mentioned above, the goal of RCS is to dynamically allocate smaller weights to tasks that are converging more rapidly, and larger weights to those converging more slowly, such that all tasks can converge at the same time. This score is calculated based on the convergence slopes of all tasks at a specific iteration $t$, that is,
\begin{equation} 
\RCS_i(t) = \softmax_i\bigg(\frac{K \alpha_{i}(t)}{\sum_{i=1}^{K} \lvert \alpha_{i}(t) \rvert}\bigg), \label{eqa:RCS} 
\end{equation} 
where $\softmax_i$ means that the softmax operation is applied to the dimension of $i$ (i.e., the dimension of tasks). To guarantee a level playing field across all tasks, we first normalize the convergence slopes as $\alpha_{i}(t) / \sum_{i=1}^K |\alpha_{i}(t)|$, making the calculated score resistant to variations in the mean scale of the slopes. However, given that this normalized value tends towards zero as the number of tasks $K$ increases, we compensate by the multiplication of $K$. This adjustment ensures that the final RCSs are not disproportionately affected by the total number of tasks being considered. The subsequent application of the softmax function can then effectively differentiate the RCS values across the tasks.

In practice, as displayed in Figure~\ref{fig1-a}, Task B, highlighted by the blue curve, demonstrates the slowest convergence rate, which is appropriately reflected by the highest RCS, depicted in Figure~\ref{fig1-c}. Additionally, Figure~\ref{fig1-a} shows that Task C (the red curve) converges slower than Task A (the green curve) up to step 8000, which is coherently translated into a higher RCS for Task C compared to Task A when $t \leq 8000$ (cf. Figure~\ref{fig1-c}).

\subsection{Absolute Convergence Scores (ACS)}
Unfortunately, relying solely on RCS proves to be inadequate for the needs of multi-task learning. As depicted in Figures~\ref{fig1-a} and~\ref{fig1-c}, Task B illustrates a scenario where, despite beginning to diverge, it still secures the highest RCS due to its largest (albeit positive) convergence slope. Awarding the greatest weight to Task B under these circumstances could exacerbate the situation by leading to further overfitting on this task, potentially causing overall model performance to deteriorate—a scenario we intend to avoid. This predicament underscores the necessity of ACS, whose fundamental purpose is to mitigate such risks by allocating reduced weights to tasks that are diverging, while favoring tasks that are on a converging trajectory with larger weights. The ACS for a given task $i$ at any step $t$ is mathematically represented as:
\begin{equation}
\ACS_i(t) = \softmax_i\bigg(\frac{-N\alpha_i(t)}{\sum_{j=t-N+1}^t \lvert \alpha_i(j) \rvert}\bigg).
\label{eqa:ACS}
\end{equation}
Unlike RCS, where both normalization within the softmax and the softmax itself occur across the task dimension $i$, ACS performs normalization along the iteration dimension $t$ from step $t-N+1$ to step $t$, but subsequently applies the softmax function across the task dimension $i$. ACS's unique aspect lies in its exclusive consideration of a task's own historical performance during the normalization, without considering other tasks. This isolation of individual task trajectory is the reason behind the nomenclature ``Absolute'' in ACS.

In general, in the initial stages of fine-tuning, tasks typically exhibit fast convergence, marked by a substantial negative slope. For tasks that maintain a consistent convergence, these negative slopes exhibit minimal change over the span from $t-N+1$ to $t$. In contrast, tasks that start to diverge will show significant changes in their slopes, transitioning from negative to neutral or even positive values. By normalizing the slope over this time window and accounting for the negative sign as shown in Eq.~\eqref{eqa:ACS}, tasks that continue to converge receive relatively high values at step $t$. Conversely, tasks that begin to diverge are assigned progressively lower values. The subsequent application of the softmax function across tasks allows us to allocate weights appropriately, thereby achieving the desired effect of bolstering converging tasks and restraining the influence of diverging ones.

Figure \ref{fig1-a}, \ref{fig1-b} and \ref{fig1-d} intuitively demonstrate the utility of the ACSs. As Task B's loss ratio diverges at the earliest, its convergence slope rapidly approaches zero, resulting in the smallest ACS. Additionally, before step 5900, Task C's convergence slope approaches zero faster than Task A's, thereby receiving a lower ACS. After 5900 steps, the convergence slope value of Task A exceeds Task C's, indicating that Task A will diverge earlier than Task C, thus a lower ACS is attributed to Task A.

\begin{algorithm}[t]
    \caption{CoBa}
    \label{algo:1}
    \renewcommand{\algorithmicrequire}{\textbf{Input:}}
    \renewcommand{\algorithmicensure}{\textbf{Output:}}
    \begin{algorithmic}[1]\small
        \REQUIRE Initial parameter $\theta_{0}$, $M$ batches of validation set, history window length $N=5M$, warm-up steps $W=M$, task number $K$, $\omega_{i}(0)=1/K$, validation loss ratios window $\vy_{i}(N;0) \leftarrow []$
        \ENSURE Trained parameter $\theta$
        \FOR{$t=0:T$}
            \STATE Compute $\ell(\theta;t)$ with training batch $\vx_{t}$
            \STATE Compute $\Bar{\ell}_i^{val}(\theta;t)$ with validation batch $\vv_{t}$
            \STATE $\vy_{i}(N;t) \leftarrow [\Bar{\ell}_i^{val}(\theta;s_0),...,\Bar{\ell}_i^{val}(\theta;t)]^\top$
            \STATE Compute $\alpha_{i}(t)$ with (\ref{eqa:6}),(\ref{eqa:7})
            \IF{$t \le W$}
                \STATE Compute $RCS(t)$ with (\ref{eqa:RCS})
                \STATE Compute $ACS(t)$ with (\ref{eqa:ACS})
                \STATE Compute $DF(t)$ with (\ref{eqa:16})
                \STATE Compute $\omega(t)$ with (\ref{eqa:19})
            \ELSE
                \STATE Compute $\omega_{i}(t) = \frac{1}{K}$
            \ENDIF
        \ENDFOR
    \end{algorithmic}
\end{algorithm}

\subsection{Divergence Factor and Final Weight}
In practice, it's common that at the onset of training, tasks generally show converging patterns. Consequently, during this phase, RCS should play the primary role in dictating the weights assigned to each task's loss. Nevertheless, as training progresses, it may happen that some tasks begin to diverge. In such instances, ACS ought to take precedence in influencing the task loss weights. To seamlessly transition from RCS-dominance to ACS-dominance in response to these evolving conditions, we introduce the concept of a divergence factor (DF), designed to monitor divergence trends throughout the training process. Given the divergence factor $\DF(t)$ at step $t$, we can compute the final weight vector $\vomega(t) = [\omega_1(t),\cdots,\omega_K(t)]^\top$ that takes both RCS and ACS into account as:
\begin{equation}
\vomega(t) = \DF(t) \RCS(t) + (1-\DF(t)) \ACS(t).
\label{eqa:19}
\end{equation}
Now let us delve into the calculation of $\DF(t)$. The approach to determining the DF involves capturing the largest (considering signs) convergence slope, denoted as $\alpha_{\max}(t)$, across all tasks at each iteration $t$.
The DF itself is then quantified by the formula:
\begin{equation}
\DF(t) \! = \! \min \! \bigg( \! t\, \softmax_t\Big(- \frac{\tau t\alpha_{\max} (t)}{\sum_{i=1}^t  \alpha_{\max} (i)} \Big), 1 \! \bigg),
\label{eqa:16}
\end{equation}
where $\softmax_t$ denotes that the softmax operation is applied to the dimension of steps. Crucially, within the softmax function, we multiply the numerator by the current step $t$ to ensure that $\DF(t)$ does not inherently decline as $t$ increases—even though the denominator naturally accumulates over time. The use of the temperature parameter $\tau > 1$ assures a sufficiently high level of distinction between the softmax outputs. Finally, the entire $\softmax$ term is scaled by $t$ to guarantee that $\DF(t)$ equals 1 when all tasks are continuously converging. Concretely, suppose that the most slowly converging task sustains a constant negative slope. $\softmax_t$ then yields $1/t$ at step $t$, suggesting a decreasing proportion of RCS in the final weight as training proceeds. However, in this context where all tasks are converging, RCS should retain its dominance in the final weight. Thus, by multiplying the $\softmax_t$ outputs by $t$, we ensure $\DF(t)$ remains at 1. On the other hand, $\DF(t)$ given by Eq.~\eqref{eqa:16} falls below 1 only when the slope of a task keeps increasing from negative to zero or even positive—indicative of the onset of divergence, which aligns with the intended design of our method.

Illustrating the proposed method with the example provided in Figure~\ref{fig1}, we note that in the initial training stages—say, before 700 steps—the gradual tapering of the DF (see Figure~\ref{fig1-e}) allows RCS to exert a stronger influence, leading to Task B receiving the heaviest weight. However, as Task B's convergence slope swiftly nears zero, the DF undergoes a swift decline, hence amplifying the role of ACS. Our methodology adeptly captures the point at which the convergence slope of Task B starts oscillating around zero, resulting in a lower ACS for Task B. Post the 700-step mark, Task B's weight is reduced significantly, a strategic move to effectively mitigate the risk of overfitting.

\vspace{-1ex}
\paragraph{Difference from Existing Methods:}~Current approaches to convergence balancing, such as GradNorm~\cite{chen2018gradnorm}, DWA~\cite{liu2019end}, LBTW~\cite{liu2019loss}, FAMO~\cite{liu2024famo}, and MetaWeighting~\cite{mao2022metaweighting}, are designed around the first criterion \textbf{\textit{c1}} outlined at the beginning of this section: decelerating the convergence of rapidly converging tasks while accelerating the convergence of slower tasks. The proposed RCS also accomplishes this objective effectively. Yet, it should be noted that this first criterion often has a counteractive effect on convergence balancing when certain tasks start to diverge. This is an issue that existing methods fail to address. To counteract this, CoBa introduces the ACS, which assigns lower weights to tasks that are diverging. Furthermore, DF improves this by detecting the divergence trend of tasks and subsequently magnifying the importance of ACS. This suppresses premature divergence trends, ensuring overall stability is maintained.
\vspace{-1ex}

\subsection{Complexity Analysis}
The overall algorithm is summarized in Algorithm~\ref{algo:1}. Here, we provide an analysis of CoBa's computational complexity. For our assumptions, we assign the computational complexity of forward propagation as $F$ and the complexity of backward propagation as $B$. We denote the number of tasks as $K$, the length of the history window as $N$, and the number of training iterations as $T$. Initially, CoBa calculates the loss for a training batch which updates the parameters, and this process costs $\gO(F+B)$ time. Subsequently, it evaluates the validation batch's loss, taking $\gO(F)$ time. Then, it calculates the convergence slopes $\valpha(t) = [\alpha_1(t),\cdots,\alpha_K(t)]^\top$ which requires $\gO(2KN)$ flops. The computation of $\RCS(t)$ and $\ACS(t)$ costs $\gO(5K)$ and $\gO(3K+2N)$ time, respectively. Ultimately, the identification of $\DF(t)$ and weight consumes $\gO(7T)$ and $\gO(3K)$ time, respectively. Thus, the joint time complexity of CoBa is $\gO(2F+B+2KN+11K+2N+7T)$. In terms of $F$, $B$, and $K$, this expression can be simplified to $\gO(2F+B+a_{3}K)$, as shown in Table~\ref{tab:time-complexity}.

%

\section{Experiments}
\label{sec:experiment}

In this section, we assess the performance of the CoBa across four diverse datasets: the \textbf{Code Completion (CC) Dataset}, encompassing five programming languages; the \textbf{Code-Related Task (CRT) Dataset}, featuring five unique programming tasks; the \textbf{XTREME-UP}, which delves into question-answering across nine natural languages; and \textbf{Multi-Domain QA Dataset}, including question answering data in the fields of coding, mathematics, and natural language. Due to the space limit, the results of the last dataset are shown in Appendix~\ref{app:results_multi_domain_qa}. The tasks within these datasets are inherently related and generative, making them ideal candidates for MTL experiments on LLM. For further insights into the datasets, readers are directed to the Appendix~\ref{app:datasets}.

\begin{figure*}[t]
    \centering
    \includegraphics[width=0.999\textwidth]{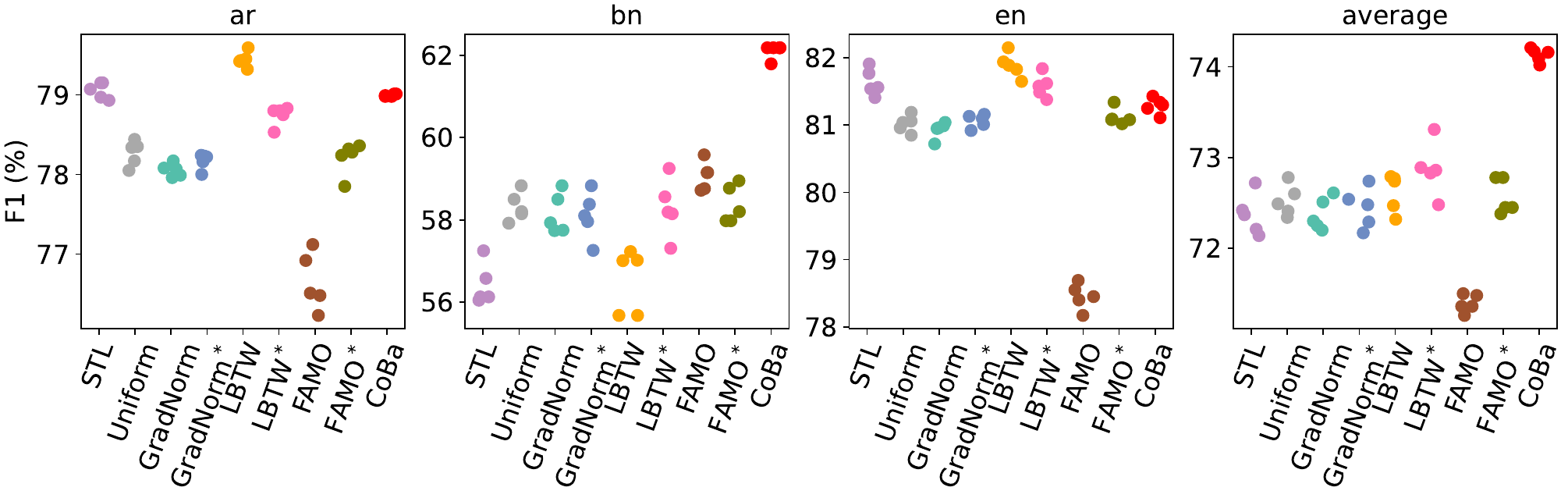}
    \vspace{-3ex}
    \caption{Experimental results on XTREME-UP dataset with 3-tasks setting.}
    \label{fig:xtreme-up-3tasks}
\end{figure*}

Our evaluation benchmarks the CoBa against 8 state-of-the-art (SOTA) baselines\footnote{We exclude MetaWeighting due to its high computational demands, as detailed in Table~\ref{tab:time-complexity}, thereby rendering it prohibitive for use with LLMs.}: \textbf{Single-Task Learning (STL)}, which finetunes each task in isolation; \textbf{Uniform}~\cite{liu2023mftcoder}, applying equal weights to all tasks in an MTL framework; \textbf{GradNorm}~\cite{chen2018gradnorm}, a method that optimizes the task weights iteratively such that task-specific gradients are of similar magnitude; \textbf{LBTW}~\cite{liu2019loss}, which dynamically adjusts task weights according to the ratio of current to initial loss $\omega_i(t) = (\ell_i(t) / \ell_0(t))^b$, parameterized by a hyperparameter $b$; and \textbf{FAMO}~\cite{liu2024famo}, aimed at optimizing weights to enhance the minimal improvement rate across tasks. Notably, the last three methods were originally designed based on the training loss. In pursuit of enhanced generalization for the fine-tuned models, we have adapted these methods to focus on validation loss, denoted as \textbf{LBTW$^*$}, \textbf{GradNorm$^*$}, and \textbf{FAMO$^*$}. Except for STL and Uniform, all methods strive to balance convergence across tasks, demonstrating their potential to compete with CoBa. The detailed experiment setup is described in Appendix~\ref{app:setup}.

\begin{table}[t]
    \centering
    \caption{Performance on the CC Dataset with Phi-1.5-1.3B model.}
    \vspace{-1ex}
    \resizebox{\linewidth}{!}{
    \begin{tabular}{lcccccc}
        \toprule
        \pmb{Method} & \pmb{Python} & \pmb{Java} & \pmb{C++} & \pmb{JS} & \pmb{Go} & \pmb{Avg} \\
        \midrule
        STL & 48.8 & 23.8 & 20.7 & 26.8 & 17.1 & 27.4 \\
        Uniform & 48.2 & 22.6 & 23.2 & 27.4 & 16.5 & 27.6 \\
        GradNorm & 47.0 & 22.0 & 23.2 & 26.8 &  16.5 & 27.1 \\
        GradNorm$^{*}$ & 47.6 & 25.6 &  22.0 & 26.2 & 17.1 & 27.7 \\
        LBTW & 47.0 & 22.0 & 22.3 & 28.1 & \pmb{18.3} & 27.6 \\
        LBTW$^{*}$ & 48.2 & 23.2 & 23.8 & 26.8 & \pmb{18.3} & 28.1 \\
        FAMO & 48.2 & \pmb{26.2} & 21.3 &   26.2 &  17.1 & 27.8 \\
        FAMO$^{*}$ & 48.2 & 23.2 & \pmb{25.6} & 26.8 & 17.7 & 28.3 \\
        CoBa & \pmb{49.4} & 24.4 & 25.0 & \pmb{29.9} & \pmb{18.3} & \pmb{29.4} \\
        \bottomrule
    \end{tabular}}
    \label{tab:5languages-phi-result}
\end{table}

\subsection{Results for CC and CRT}
Table~\ref{tab:5languages-phi-result} shows the Pass@1 metric for all code completion (CC) tasks resulting from all methods. Moreover, Figure~\ref{fig:5languages-loss} graphically presents the normalized validation loss ratio across all tasks for each method. \textbf{CoBa demonstrates superior performance over the baseline methods in the Pass@1 metric for five programming languages, achieving a minimum of 4\% relative improvement in the average Pass@1 score} (calculated as $(29.4 - 28.3) / 28.3 = 4\%$). In addition, adaptations of FAMO, LBTW, and GradNorm to the validation loss rather than the training loss (i.e., FAMO$^*$, LBTW$^*$, and GradNorm$^*$) show enhanced performance. This enhancement signifies the importance of balancing convergence speed based on validation loss for better generalization, as pointed out in~\cite{mao2022metaweighting}. Indeed, FAMO$^*$ closely trails the performance of CoBa. However, Figure~\ref{fig:5languages-loss} reveals its limitation in preemptively addressing the divergence in the Python completion task, thus limiting its overall efficacy. As an alternative, by utilizing the ACS and DF, CoBa effectively neutralizes the divergence of the Python task. Contrastingly, despite its aims of learning all tasks at an equal pace, GradNorm's performance lags behind counterparts such as CoBa, FAMO, and LBTW. This underperformance may be attributed to GradNorm's strategy of adjusting loss weights using the same learning rate as the model parameters, a tactic that proves ineffective due to the typically small learning rates employed in training LLMs. Consequently, the weights adjusted by GradNorm remain almost identical to the initial uniform weights, failing to dynamically respond to the learning progress and hampering convergence balance. 

\begin{table}[t] 
    \centering
    \caption{Performance on the CC Dataset with CodeLlama-13B-Python.}
    \label{tab:5languages-codellama-result}
    \vspace{-1ex}
    \resizebox{\linewidth}{!}{
    \begin{tabular}{lcccccc}
        \toprule
        \pmb{Method} & \pmb{Python} & \pmb{Java} & \pmb{C++} & \pmb{JS} & \pmb{Go} & \pmb{Avg} \\
        \midrule
        STL & 60.4 & 53.7 & 40.2 & 53.7 & 37.2 & 49.0 \\
        Uniform & 62.2 & 53.1 & 40.8 & 51.8 & 39.6 & 49.5 \\
        GradNorm & 62.8 & 53.1 & 41.5 & 52.4 & 37.2 & 49.4 \\
        GradNorm$^{*}$ & 61.0 & 54.3 & 40.9 & 53.7 & 39.6 & 49.9 \\
        LBTW & 64.0 & 54.9 & 40.8 & 51.8 & 39.0 & 50.1 \\
        LBTW$^{*}$ & 62.2 & 52.4 & \pmb{42.7} & 54.9 & 40.8 & 50.6 \\
        FAMO & 62.2 & 52.4 & 42.1 & 51.8 & 41.5 & 50.0 \\
        FAMO$^{*}$ & 62.2 & 53.1 & 41.5 & 53.1 & 41.5 & 50.2 \\
        CoBa & \pmb{65.9} & \pmb{56.7} & \pmb{42.7} & \pmb{56.7} & \pmb{42.7} & \pmb{52.9} \\
        \bottomrule
    \end{tabular}}
    \vspace{-3ex}
\end{table}

Regarding the CRT Dataset, its results are similar to those of the CC dataset, and so we defer its detailed discussion to Appendix~\ref{app:results_code_related}. Notably, in contrast with other state-of-the-art methods, \textbf{CoBa excelled in the Code Completion and Unit Test Generation tasks, recording substantial relative average Pass@1 improvements of at least 6\% and 13\%, respectively.}


\subsection{Results for XTREME-UP}
In this study, we conduct experiments across three groups, each consisting of 3, 6, and 9 tasks, with a mix of high and low-resource languages. We perform five trials per group to assess the resilience of our proposed method, CoBa, against varying task quantities and its capability to generalize performance for low-resource languages. The results, presented in Figure~\ref{fig:xtreme-up-3tasks} and Figures~\ref{fig:xtreme-up-6tasks} and~\ref{fig:xtreme-up-9tasks} in the appendix, consistently show \textbf{CoBa outperforming all baselines in terms of the average span $F_1$ score across all conditions}. Notably, the effectiveness of CoBa remains stable regardless of the number of tasks, illustrating its adaptability. Importantly, CoBa showcases pronounced enhancements in performance for low-resource languages, like Bengali (bn) and Telugu (te), with a 3\% to 5\% absolute increase in span $F_1$ scores over the Single-Task Learning (STL) approach. This underscores CoBa's proficiency in improving generalization for tasks with limited data availability. For high-resource languages, CoBa's performance matches or surpasses that of STL, suggesting that balancing convergence can catalyze synergistic benefits among related tasks. Our experiment also reveals that FAMO generally underperforms, likely due to its sensitivity to the regularization coefficient $\gamma$~\cite{liu2024famo}, which requires manual customization for each dataset. 
In contrast, FAMO$^*$, designed for the validation set, bypasses re-normalization and shows much better performance.\footnote{
We have also identified several other factors that may help explain the gap between FAMO and FAMO$^*$:
\begin{enumerate}[leftmargin=*, topsep=1pt,itemsep=1pt,partopsep=1pt,parsep=1pt]
    \item 
    Utilizing the convergence properties of the validation set, rather than the training set, for task weight allocation can lead to improved performance.
    \item FAMO optimizes an approximation of the original loss to facilitate the reuse of intermediate computations for weight updates, thereby reducing computational complexity. However, this approximation is only effective with an appropriately set learning rate; otherwise, it can create a performance gap. In contrast, FAMO$^*$ optimizes the original training loss without performing re-normalization. Indeed, we observe that FAMO exhibits a higher loss scale (both training and validation) in nearly all experiments compared to FAMO$^*$, except in the Multi-Domain QA dataset where a higher learning rate is applied. This disparity is particularly pronounced in the XTREME-UP experiments, where the loss for FAMO is approximately double that of FAMO$^*$.
    \item The F1 score metric employed for XTREME-UP has a strong correlation with the loss, which elucidates why FAMO's performance on this dataset is significantly inferior to that of FAMO$^*$. Conversely, for code-related datasets, the Pass@1 metric shows a relatively weak correlation with loss; thus, a decrease in loss does not necessarily translate to an increase in Pass@1. This may account for the comparable performance of FAMO and FAMO$^*$ in these code-related datasets.
\end{enumerate}}

\subsection{Ablation Study and Run Time Analysis}
\begin{figure*}[t]
    \centering
    \includegraphics[width=0.9\textwidth]{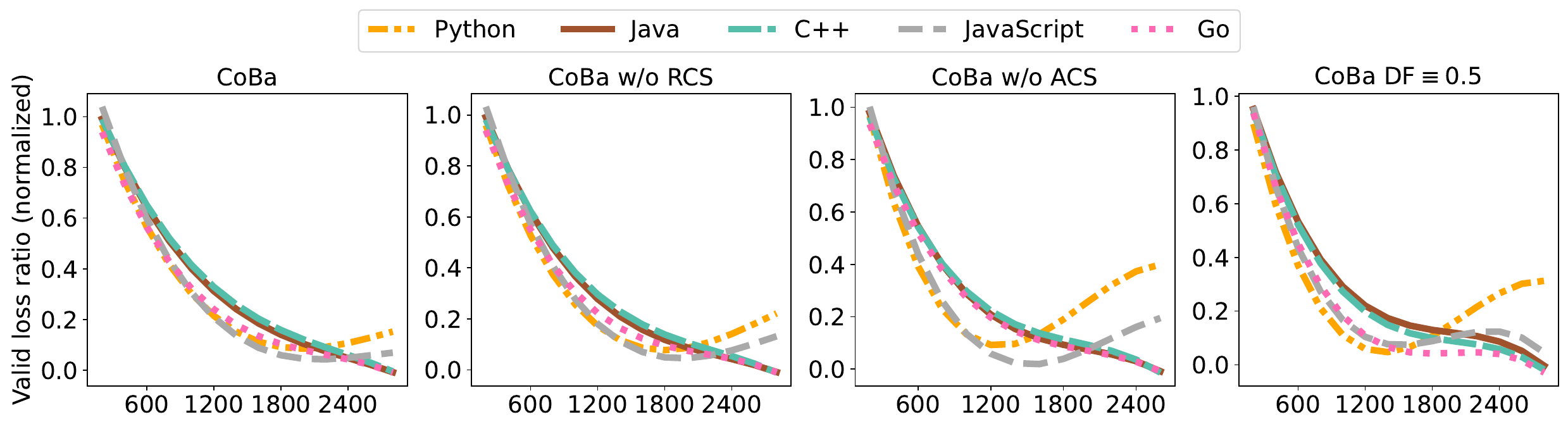}
    \vspace{-1ex}
    \caption{Normalized valid loss ratio of ablation study on the CC dataset for 5 programming languages. For better visualization, we apply Min-Max Normalization to the validation loss ratios for each task.}
    \label{fig:5languages-loss-ablation}
    \vspace{-0.5ex}
\end{figure*}

We first examine \textbf{the impact of RCS, ACS, and DF} within CoBa.
Figure~\ref{fig:5languages-loss-ablation} highlights the necessity of combining all three components to ensure that all tasks converge at a similar pace. Moreover, the quantitative results in Table~\ref{tab:5languages-phi-ablation} reveal a decline in CoBa’s effectiveness when excluding either RCS, ACS, or DF. Next, we evaluate \textbf{CoBa’s adaptability across models of varying sizes} by choosing CodeLlama-13B-Python as the base model. Results in Table~\ref{tab:5languages-codellama-result} highlight CoBa's exceptional performance across all five programming languages, boasting a minimum of a 5\% enhancement in average Pass@1 relative to other SOTA methods. Comparisons between the larger CodeLlama-13B-Python and the smaller Phi-1.5-1.3B models—seen in Table~\ref{tab:5languages-phi-result}—highlighted that larger models boost CoBa's multi-task learning efficacy. This suggests CoBa's compatibility with and enhanced performance through the utilization of larger models. Finally, we analyze the \textbf{runtime efficiency} of CoBa on both the CodeLlama-13B-Python and Phi-1.5-1.3B models, in comparison to other methods. As expected from our theoretical analysis (see Table~\ref{tab:time-complexity}), CoBa requires a significantly shorter runtime than other validation set-based convergence balancing methods like GradNorm$^*$ and FAMO$^*$, and aligns closely with Uniform, the most straightforward MTL approach. This efficiency further positions CoBa as a practical choice for integrating into MTL frameworks for LLMs. 

\begin{table}[t]
    \centering
    \caption{Comparison of the time taken per epoch for experiments on the CC Dataset.}
    \vspace{-1ex}
    \resizebox{0.95\linewidth}{!}{
    \begin{tabular}{lcc}
        \toprule
        \multirow{2}{*}{\pmb{Method}} & \pmb{Phi-1.5-1.3B} & \pmb{CodeLlama-13B-Python} \\
        & \pmb{mins/epoch} & \pmb{mins/epoch} \\
        \midrule
        Uniform & 22.98 & 188.93 \\
        GradNorm & 25.80 & 197.10 \\
        GradNorm$^{*}$ & 46.08 & 351.13 \\
        LBTW & 26.15 & 195.28 \\
        LBTW$^{*}$ & 29.20 & 234.77 \\
        FAMO & 24.83 & 192.53 \\
        FAMO$^{*}$ & 41.58 & 335.82 \\
        CoBa & 29.05 & 230.20 \\
        \bottomrule
    \end{tabular}}
    \label{tab:5languages-time}
    \vspace{-0.5ex}
\end{table}

\section{Conclusion}
\label{sec:conclusion}

In this paper, we propose CoBa, a novel MTL method for LLMs that simultaneously achieves convergence balance with low computational complexity. Extensive experiments on four real-world datasets have demonstrated the efficacy and efficiency of the proposed method.

\section{Ethical Considerations}
Our research is foundational and not expected to have significant social implications. We ensure transparency and adherence to ethical standards in the use of datasets. Additionally, the accessibility of these datasets is beneficial for broader reproducibility and review within the research community, aligning with ethical research practices. However, we acknowledge the responsibility that comes with the development of any MTL technology. We encourage ongoing dialogue and ethical considerations in the application of our findings.

\section{Limitations}
We have identified the main limitations of our approach as follows:
\begin{itemize}[leftmargin=*, topsep=1pt,itemsep=1pt,partopsep=1pt,parsep=1pt]
    \item \textbf{The Model Parameter Scale:} Due to resource constraints, we are unable to evaluate the efficacy of CoBa on larger LLMs. In future work, we aspire to conduct experiments with larger LLMs, akin to MFTCoder~\cite{liu2023mftcoder}, to further substantiate our findings.
    \item \textbf{The Number of Tasks:} Due to the limited number of open-source multi-task fine-tuning data, we are unable to experiment on more tasks. In the future, we hope to collect more relevant multi-task datasets to verify the effectiveness of CoBa.
    \item \textbf{Domain of application:} This paper focuses on NLP. However, multi-task learning is not limited to this modality. In the future, we aim to explore other modalities such as computer vision.
    \item \textbf{Task Conflicts or Interference:} CoBa is designed to achieve convergence balance among tasks and does not guarantee optimal performance for all tasks in the presence of conflicts. A promising solution is to integrate CoBa with a Mixture of Experts (MoE) framework, assigning each task to a specific expert within the model. This separation enables tasks to have individualized sets of parameters, mitigating the issue of task interference.
    \item \textbf{Curriculum Learning:} While CoBa prioritizes difficult tasks at the initial stage of Multi-Task Learning (MTL), curriculum learning emphasizes prioritizing easier tasks, which can be advantageous in scenarios where learning harder tasks may become easier once the model has mastered the easier tasks. Therefore, the first criterion of CoBa may not be directly applicable in this setup. Nonetheless, an interesting modification to CoBa could be its ability to automatically identify and assign greater weight to easier tasks (e.g., tasks that converge faster) during the initial training stages to align with curriculum learning principles. Furthermore, the debate on whether to prioritize easy tasks over difficult ones or vice versa, as noted in~\cite{guo2018dynamic}, is also an ongoing and important research topic. It is, therefore, worth exploring how to modify CoBa to dynamically decide which tasks to focus on during different training stages, ensuring that all tasks are well learned in the final stage.
    
\end{itemize}

\bibliography{custom}

\begin{thebibliography}{31}
\expandafter\ifx\csname natexlab\endcsname\relax\def\natexlab#1{#1}\fi

\bibitem[{Achiam et~al.(2023)Achiam, Adler, Agarwal, Ahmad, Akkaya, Aleman, Almeida, Altenschmidt, Altman, Anadkat et~al.}]{achiam2023gpt}
Josh Achiam, Steven Adler, Sandhini Agarwal, Lama Ahmad, Ilge Akkaya, Florencia~Leoni Aleman, Diogo Almeida, Janko Altenschmidt, Sam Altman, Shyamal Anadkat, et~al. 2023.
\newblock Gpt-4 technical report.
\newblock \emph{arXiv preprint arXiv:2303.08774}.

\bibitem[{Aghajanyan et~al.(2021)Aghajanyan, Gupta, Shrivastava, Chen, Zettlemoyer, and Gupta}]{aghajanyan2021muppet}
Armen Aghajanyan, Anchit Gupta, Akshat Shrivastava, Xilun Chen, Luke Zettlemoyer, and Sonal Gupta. 2021.
\newblock Muppet: Massive multi-task representations with pre-finetuning.
\newblock In \emph{Proceedings of the 2021 Conference on Empirical Methods in Natural Language Processing}, pages 5799--5811.

\bibitem[{Aribandi et~al.(2021)Aribandi, Tay, Schuster, Rao, Zheng, Mehta, Zhuang, Tran, Bahri, Ni et~al.}]{aribandi2021ext5}
Vamsi Aribandi, Yi~Tay, Tal Schuster, Jinfeng Rao, Huaixiu~Steven Zheng, Sanket~Vaibhav Mehta, Honglei Zhuang, Vinh~Q Tran, Dara Bahri, Jianmo Ni, et~al. 2021.
\newblock Ext5: Towards extreme multi-task scaling for transfer learning.
\newblock In \emph{International Conference on Learning Representations}.

\bibitem[{Bai et~al.(2023)Bai, Bai, Chu, Cui, Dang, Deng, Fan, Ge, Han, Huang et~al.}]{bai2023qwen}
Jinze Bai, Shuai Bai, Yunfei Chu, Zeyu Cui, Kai Dang, Xiaodong Deng, Yang Fan, Wenbin Ge, Yu~Han, Fei Huang, et~al. 2023.
\newblock Qwen technical report.
\newblock \emph{arXiv preprint arXiv:2309.16609}.

\bibitem[{Chen et~al.(2021)Chen, Tworek, Jun, Yuan, Pinto, Kaplan, Edwards, Burda, Joseph, Brockman et~al.}]{chen2021evaluating}
Mark Chen, Jerry Tworek, Heewoo Jun, Qiming Yuan, Henrique Ponde de~Oliveira Pinto, Jared Kaplan, Harri Edwards, Yuri Burda, Nicholas Joseph, Greg Brockman, et~al. 2021.
\newblock Evaluating large language models trained on code.
\newblock \emph{arXiv preprint arXiv:2107.03374}.

\bibitem[{Chen et~al.(2018)Chen, Badrinarayanan, Lee, and Rabinovich}]{chen2018gradnorm}
Zhao Chen, Vijay Badrinarayanan, Chen-Yu Lee, and Andrew Rabinovich. 2018.
\newblock Gradnorm: Gradient normalization for adaptive loss balancing in deep multitask networks.
\newblock In \emph{International conference on machine learning}, pages 794--803. PMLR.

\bibitem[{Chen et~al.(2020)Chen, Ngiam, Huang, Luong, Kretzschmar, Chai, and Anguelov}]{chen2020just}
Zhao Chen, Jiquan Ngiam, Yanping Huang, Thang Luong, Henrik Kretzschmar, Yuning Chai, and Dragomir Anguelov. 2020.
\newblock Just pick a sign: Optimizing deep multitask models with gradient sign dropout.
\newblock \emph{Advances in Neural Information Processing Systems}, 33:2039--2050.

\bibitem[{Crawshaw(2020)}]{crawshaw2020multi}
Michael Crawshaw. 2020.
\newblock Multi-task learning with deep neural networks: A survey.
\newblock \emph{arXiv preprint arXiv:2009.09796}.

\bibitem[{Di et~al.(2024)Di, Li, Yu, Jiang, Cai, Cao, Chen, Chen, Chen, Chen et~al.}]{di2023codefuse}
Peng Di, Jianguo Li, Hang Yu, Wei Jiang, Wenting Cai, Yang Cao, Chaoyu Chen, Dajun Chen, Hongwei Chen, Liang Chen, et~al. 2024.
\newblock Codefuse-13b: A pretrained multi-lingual code large language model.
\newblock In \emph{Proceedings of the 46th International Conference on Software Engineering: Software Engineering in Practice}, pages 418--429.

\bibitem[{Guo et~al.(2018)Guo, Haque, Huang, Yeung, and Fei-Fei}]{guo2018dynamic}
Michelle Guo, Albert Haque, De-An Huang, Serena Yeung, and Li~Fei-Fei. 2018.
\newblock Dynamic task prioritization for multitask learning.
\newblock In \emph{Proceedings of the European conference on computer vision (ECCV)}, pages 270--287.

\bibitem[{Kendall et~al.(2018)Kendall, Gal, and Cipolla}]{kendall2018multi}
Alex Kendall, Yarin Gal, and Roberto Cipolla. 2018.
\newblock Multi-task learning using uncertainty to weigh losses for scene geometry and semantics.
\newblock In \emph{Proceedings of the IEEE conference on computer vision and pattern recognition}, pages 7482--7491.

\bibitem[{Li et~al.(2023)Li, Bubeck, Eldan, Del~Giorno, Gunasekar, and Lee}]{li2023textbooks}
Yuanzhi Li, S{\'e}bastien Bubeck, Ronen Eldan, Allie Del~Giorno, Suriya Gunasekar, and Yin~Tat Lee. 2023.
\newblock Textbooks are all you need ii: phi-1.5 technical report.
\newblock \emph{arXiv preprint arXiv:2309.05463}.

\bibitem[{Lin et~al.(2021)Lin, Feiyang, and Zhang}]{lin2021closer}
Baijiong Lin, YE~Feiyang, and Yu~Zhang. 2021.
\newblock A closer look at loss weighting in multi-task learning.

\bibitem[{Liu et~al.(2024{\natexlab{a}})Liu, Chen, Liao, Gong, Wang, Lei, Liang, Chen, Shen, Zhou et~al.}]{liu2023mftcoder}
Bingchang Liu, Chaoyu Chen, Cong Liao, Zi~Gong, Huan Wang, Zhichao Lei, Ming Liang, Dajun Chen, Min Shen, Hailian Zhou, et~al. 2024{\natexlab{a}}.
\newblock Mftcoder: Boosting code llms with multitask fine-tuning.
\newblock In \emph{Proceedings of the 30th ACM SIGKDD Conference on Knowledge Discovery and Data Mining}.

\bibitem[{Liu et~al.(2024{\natexlab{b}})Liu, Feng, Stone, and Liu}]{liu2024famo}
Bo~Liu, Yihao Feng, Peter Stone, and Qiang Liu. 2024{\natexlab{b}}.
\newblock Famo: Fast adaptive multitask optimization.
\newblock \emph{Advances in Neural Information Processing Systems}, 36.

\bibitem[{Liu et~al.(2021)Liu, Liu, Jin, Stone, and Liu}]{liu2021conflict}
Bo~Liu, Xingchao Liu, Xiaojie Jin, Peter Stone, and Qiang Liu. 2021.
\newblock Conflict-averse gradient descent for multi-task learning.
\newblock \emph{Advances in Neural Information Processing Systems}, 34:18878--18890.

\bibitem[{Liu et~al.(2020)Liu, Li, Kuang, Xue, Chen, Yang, Liao, and Zhang}]{liu2020towards}
Liyang Liu, Yi~Li, Zhanghui Kuang, Jing-Hao Xue, Yimin Chen, Wenming Yang, Qingmin Liao, and Wayne Zhang. 2020.
\newblock Towards impartial multi-task learning.
\newblock In \emph{International Conference on Learning Representations}.

\bibitem[{Liu et~al.(2019{\natexlab{a}})Liu, Liang, and Gitter}]{liu2019loss}
Shengchao Liu, Yingyu Liang, and Anthony Gitter. 2019{\natexlab{a}}.
\newblock Loss-balanced task weighting to reduce negative transfer in multi-task learning.
\newblock In \emph{Proceedings of the AAAI conference on artificial intelligence}, volume~33, pages 9977--9978.

\bibitem[{Liu et~al.(2019{\natexlab{b}})Liu, Johns, and Davison}]{liu2019end}
Shikun Liu, Edward Johns, and Andrew~J Davison. 2019{\natexlab{b}}.
\newblock End-to-end multi-task learning with attention.
\newblock In \emph{Proceedings of the IEEE/CVF conference on computer vision and pattern recognition}, pages 1871--1880.

\bibitem[{Mao et~al.(2022)Mao, Wang, Liu, Lin, and Xie}]{mao2022metaweighting}
Yuren Mao, Zekai Wang, Weiwei Liu, Xuemin Lin, and Pengtao Xie. 2022.
\newblock Metaweighting: Learning to weight tasks in multi-task learning.
\newblock In \emph{Findings of the Association for Computational Linguistics: ACL 2022}, pages 3436--3448.

\bibitem[{Mitra et~al.(2024)Mitra, Khanpour, Rosset, and Awadallah}]{mitra2024orca}
Arindam Mitra, Hamed Khanpour, Corby Rosset, and Ahmed Awadallah. 2024.
\newblock Orca-math: Unlocking the potential of slms in grade school math.
\newblock \emph{arXiv preprint arXiv:2402.14830}.

\bibitem[{Raffel et~al.(2020)Raffel, Shazeer, Roberts, Lee, Narang, Matena, Zhou, Li, and Liu}]{raffel2020exploring}
Colin Raffel, Noam Shazeer, Adam Roberts, Katherine Lee, Sharan Narang, Michael Matena, Yanqi Zhou, Wei Li, and Peter~J Liu. 2020.
\newblock Exploring the limits of transfer learning with a unified text-to-text transformer.
\newblock \emph{Journal of machine learning research}, 21(140):1--67.

\bibitem[{Roziere et~al.(2023)Roziere, Gehring, Gloeckle, Sootla, Gat, Tan, Adi, Liu, Remez, Rapin et~al.}]{roziere2023code}
Baptiste Roziere, Jonas Gehring, Fabian Gloeckle, Sten Sootla, Itai Gat, Xiaoqing~Ellen Tan, Yossi Adi, Jingyu Liu, Tal Remez, J{\'e}r{\'e}my Rapin, et~al. 2023.
\newblock Code llama: Open foundation models for code.
\newblock \emph{arXiv preprint arXiv:2308.12950}.

\bibitem[{Ruder et~al.(2023)Ruder, Clark, Gutkin, Kale, Ma, Nicosia, Rijhwani, Riley, Sarr, Wang et~al.}]{ruder2023xtreme}
Sebastian Ruder, Jonathan~H Clark, Alexander Gutkin, Mihir Kale, Min Ma, Massimo Nicosia, Shruti Rijhwani, Parker Riley, Jean Michel~Amath Sarr, Xinyi Wang, et~al. 2023.
\newblock Xtreme-up: A user-centric scarce-data benchmark for under-represented languages.
\newblock In \emph{The 2023 Conference on Empirical Methods in Natural Language Processing}.

\bibitem[{Si et~al.()Si, Wang, Gu, Liu, and Lin}]{sialpaca}
Qingyi Si, Tong Wang, Naibin Gu, Rui Liu, and Zheng Lin.
\newblock Alpaca-cot: An instruction-tuning platform with unified interface of instruction collection, parameter-efficient methods, and large language models, 2023.
\newblock \emph{URL https://github. com/PhoebusSi/alpaca-CoT}.

\bibitem[{Touvron et~al.(2023)Touvron, Martin, Stone, Albert, Almahairi, Babaei, Bashlykov, Batra, Bhargava, Bhosale et~al.}]{touvron2023llama}
Hugo Touvron, Louis Martin, Kevin Stone, Peter Albert, Amjad Almahairi, Yasmine Babaei, Nikolay Bashlykov, Soumya Batra, Prajjwal Bhargava, Shruti Bhosale, et~al. 2023.
\newblock Llama 2: Open foundation and fine-tuned chat models.
\newblock \emph{arXiv preprint arXiv:2307.09288}.

\bibitem[{Vandenhende et~al.(2021)Vandenhende, Georgoulis, Van~Gansbeke, Proesmans, Dai, and Van~Gool}]{vandenhende2021multi}
Simon Vandenhende, Stamatios Georgoulis, Wouter Van~Gansbeke, Marc Proesmans, Dengxin Dai, and Luc Van~Gool. 2021.
\newblock Multi-task learning for dense prediction tasks: A survey.
\newblock \emph{IEEE transactions on pattern analysis and machine intelligence}, 44(7):3614--3633.

\bibitem[{Xue et~al.(2023)Xue, Jain, Shah, Zheng, and You}]{xue2023instruction}
Fuzhao Xue, Kabir Jain, Mahir~Hitesh Shah, Zangwei Zheng, and Yang You. 2023.
\newblock Instruction in the wild: A user-based instruction dataset.

\bibitem[{Yu et~al.(2020)Yu, Kumar, Gupta, Levine, Hausman, and Finn}]{yu2020gradient}
Tianhe Yu, Saurabh Kumar, Abhishek Gupta, Sergey Levine, Karol Hausman, and Chelsea Finn. 2020.
\newblock Gradient surgery for multi-task learning.
\newblock \emph{Advances in Neural Information Processing Systems}, 33:5824--5836.

\bibitem[{Zhang et~al.(2023)Zhang, Yu, Yu, Guo, and Jiang}]{zhang2023survey}
Zhihan Zhang, Wenhao Yu, Mengxia Yu, Zhichun Guo, and Meng Jiang. 2023.
\newblock A survey of multi-task learning in natural language processing: Regarding task relatedness and training methods.
\newblock In \emph{Proceedings of the 17th Conference of the European Chapter of the Association for Computational Linguistics}, pages 943--956.

\bibitem[{Zheng et~al.(2023)Zheng, Xia, Zou, Dong, Wang, Xue, Wang, Shen, Wang, Li et~al.}]{zheng2023codegeex}
Qinkai Zheng, Xiao Xia, Xu~Zou, Yuxiao Dong, Shan Wang, Yufei Xue, Zihan Wang, Lei Shen, Andi Wang, Yang Li, et~al. 2023.
\newblock Codegeex: A pre-trained model for code generation with multilingual evaluations on humaneval-x.
\newblock \emph{arXiv preprint arXiv:2303.17568}.

\end{thebibliography}

\appendix

\newpage
\section{Datasets}
\label{app:datasets}

\paragraph{Code Completion (CC) Dataset}~The CC Dataset comprises five distinct programming languages: Python, Java, C++, JavaScript (JS), and Go. It is a subset derived from the code completion task data within the Code-related Tasks Dataset. Table~\ref{tab:5languages-dataset} displays the statistical information for this dataset. Training will be conducted on this dataset, with evaluations carried out on HumanEval~\cite{chen2021evaluating} and HumanEval-X~\cite{zheng2023codegeex} benchmarks, utilizing the Pass@1 metric as the assessment criterion.

\paragraph{Code-Related Task (CRT) Dataset}~The CRT Dataset~\cite{liu2023mftcoder} comprises five distinct programming tasks: code completion, code translation, Text2Code, unit testing, and code summarization. The statistical information for this dataset is presented in Table~\ref{tab:5tasks-dataset}. Evaluations for the Code Completion task will be conducted using the HumanEval~\cite{chen2021evaluating} and HumanEval-X~\cite{zheng2023codegeex} benchmarks, while the Code Translation task will be assessed using the CodeFuseEval-CodeTrans~\cite{di2023codefuse}  benchmark. The Text2Code task will utilize the MBPP for evaluation, and the Unit Test task will be evaluated using the CodeFuseEval-UnitTest~\cite{di2023codefuse} benchmark. The assessment metric for these four tasks is Pass@1. For the Code Comment task, we have constructed a test set based on 500 problems from the MBPP and will employ the BLEU score as the evaluation metric.

\paragraph{XTREME-UP}~The XTREME-UP Dataset \cite{ruder2023xtreme} is a multilingual and multitask dataset, specifically designed to address underrepresented languages in scarce-data scenarios. Our selected portion focuses on in-language question-answering sets that span across nine different languages. These languages include a mix of high-resource languages like Arabic (ar), English (en), Finnish (fi), Korean (ko), and Russian (ru), as well as low-resource languages such as Bengali (bn), Indonesian (id), Swahili (sw), and Telugu (te). A comprehensive list detailing the number of samples and data splits for each language can be found in Table~\ref{tab:xtreme-up-dataset}, as provided by~\citep{ruder2023xtreme}. It is also pertinent to mention that we have adopted the same evaluation criterion, the span $F_1$ score, as in~\citep{ruder2023xtreme}. It defines true positives as the tokens that match between the correct and generated answers. On the other hand, false positives are identified as tokens that only appear in the prediction but not in the correct answer. Lastly, tokens that are present in the correct answer but fail to appear in the prediction are classified as false negatives. Furthermore, we carry out mutually inclusive experimental groups, with three task quantities of {3, 6, 9}. Each group contains a blend of high and low-resource languages, and five trials are conducted for each to examine the resilience of our proposed method vis-a-vis the number of tasks and test the performance generalization for low-resource languages. The 3-task group is composed of Arabic (ar), Bengali (bn), and English (en), whereas the 6-task group also includes Finnish (fi), Russian (ru), and Telugu~(te).

\begin{table}[t]
    \centering
    \caption{Data statistics of the CC Dataset.}
    \resizebox{0.7\linewidth}{!}{
    \begin{tabular}{lcccccc}
        \toprule
        \pmb{Task} & \pmb{\#Samples} & \pmb{Train / Valid} \\
        \midrule
        Python & 20,539 & \multirow{5}{*}{95 / 5} \\
        Java & 32,346 & \\
        C++ & 33,291 & \\
        JavaScript & 13,217 & \\
        Go & 34,340 & \\
        \bottomrule
    \end{tabular}}
    \label{tab:5languages-dataset}
\end{table}

\begin{table}[t]
    \centering
    \caption{Data statistics of the CRT Dataset.}
    \resizebox{0.8\linewidth}{!}{
    \begin{tabular}{lcc}
        \toprule
        \pmb{Task} & \pmb{\#Samples} & \pmb{Train / Valid} \\
        \midrule
        Code Comment & 645,711 & \multirow{5}{*}{95 / 5} \\
        Code Completion & 192,547 \\
        Code Translation &  307,585 \\
        Text2Code & 94,086 \\
        Unit Test & 390,393 \\
        \bottomrule
    \end{tabular}}
    \label{tab:5tasks-dataset}
\end{table}

\begin{table}[t] 
    \centering
    \caption{Data statistics of the XTREME-UP Dataset.}
    \resizebox{\linewidth}{!}{
    \begin{small}
    \begin{tabular}{lcccc}
        \toprule
        \pmb{Task} & \pmb{\#Samples} & \pmb{Train} & \pmb{Valid} & \pmb{Test} \\
        \midrule
        Arabi (ar) & 30,401 & 26,719 & 1,841 & 1,841 \\
        Bengali (bn) & 876 & 426 & 225 & 225 \\
        English (en) & 8,121 & 6,361 & 880 & 880 \\
        Finnish (fi) & 14,676 & 11,548 & 1,564 & 1,564 \\
        Indonesian (id) & 2,684 & 426 & 1,129 & 1,129 \\
        Korean (ko) & 3,437 & 2,336 & 549 & 552 \\
        Russian (ru) & 14,140 & 10,892 & 1,624 & 1,624 \\
        Swahili (sw) & 2,387 & 425 & 965 &  997 \\
        Telugu (te) & 3,100 & 426 & 1,337 & 1,337 \\
        \bottomrule
    \end{tabular}
    \end{small}}
    \label{tab:xtreme-up-dataset}
\end{table}

\begin{table}[t]
    \centering
    \caption{Data statistics of the Multi-Domain QA Dataset.}
    \resizebox{0.7\linewidth}{!}{
    \begin{tabular}{lcc}
        \toprule
        \pmb{Task} & \pmb{\#Samples} & \pmb{Train / Valid / Test} \\
        \midrule
        Code & 94,086 & \multirow{3}{*}{95 / 5 / 5} \\
        Math & 200,035 \\
        NL &  104,133 \\
        \bottomrule
    \end{tabular}}
    \label{tab:multi-domain-qa-dataset}
\end{table}

\begin{figure*}[t]
    \centering
    \includegraphics[width=0.9\textwidth]{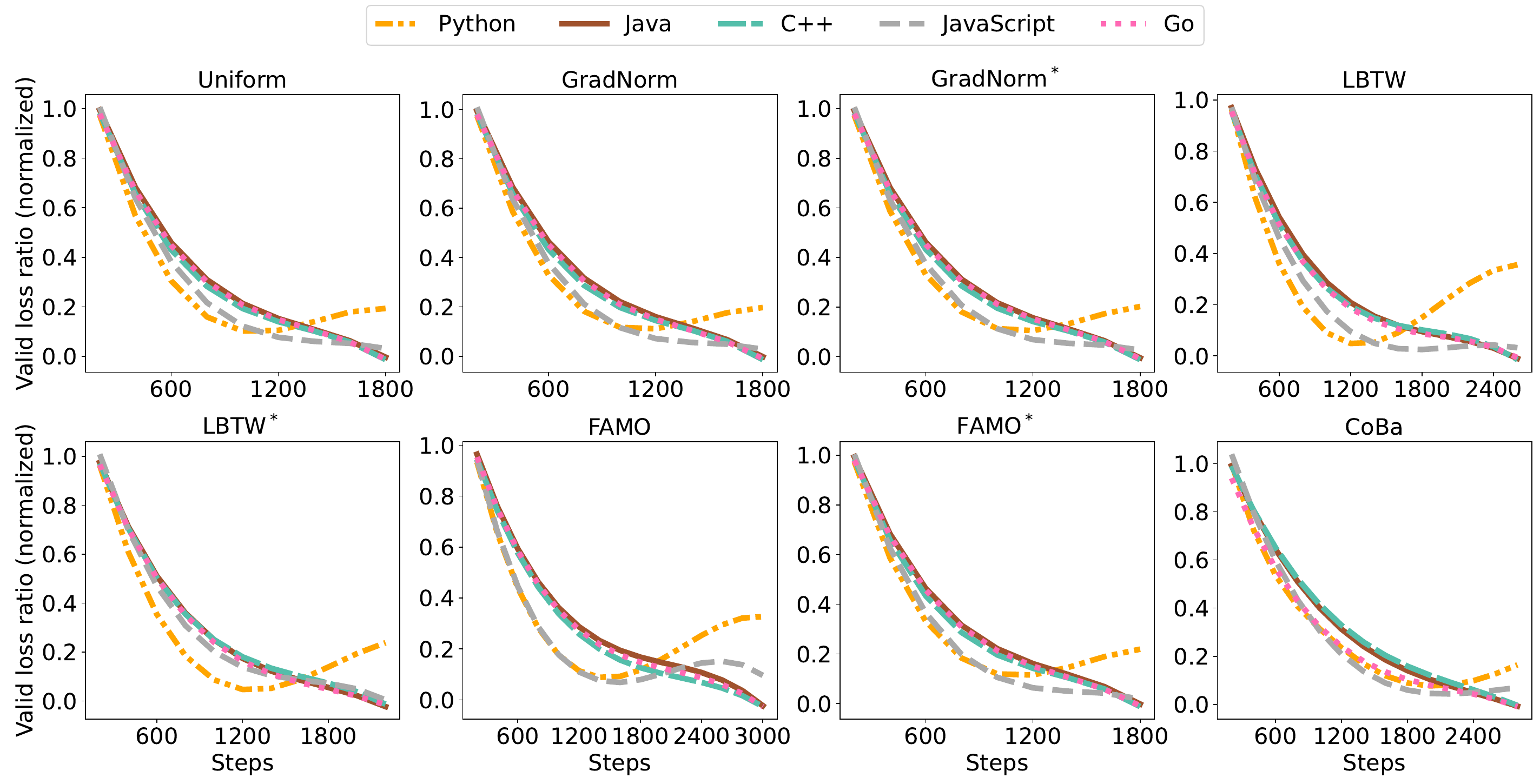}
    \caption{Normalized valid loss ratios on the CC dataset for 5 programming languages. The x-axis endpoint in each figure marks the early stopping point. For better visualization, we apply Min-Max Normalization to the validation loss ratios for each task, which involves subtracting the minimum value and then dividing by the range between the maximum and minimum values.}
    \label{fig:5languages-loss}
\end{figure*}

\begin{table}[t]
    \centering
    \caption{Ablation study on CC with Phi-1.5-1.3B.}
    \vspace{-1ex}
    \resizebox{\linewidth}{!}{
    \begin{tabular}{lcccccc}
        \toprule
        \pmb{Method} & \pmb{Python} & \pmb{Java} & \pmb{C++} & \pmb{JS} & \pmb{Go} & \pmb{Avg} \\
        \midrule
        CoBa & \pmb{49.4} & 24.4 & \pmb{25.0} & \pmb{29.9} & 18.3 & \pmb{29.4} \\
        \hspace{3mm} w/o RCS & 48.8 & 25.0 & 23.2 & 29.3 & 17.1 & 28.7 \\
        \hspace{3mm} w/o ACS & 47.0 & \pmb{26.8} & 22.6 & 26.8 & 17.1 & 28.1 \\
        \hspace{3mm} $DF \equiv 0.5$ & 47.6 & 23.8 & 23.2 & 26.8 & \pmb{20.1} & 28.3 \\
        \bottomrule
    \end{tabular}}
    \label{tab:5languages-phi-ablation}
    \vspace{-1ex}
\end{table}


\paragraph{Multi-Domain QA Dataset} The Multi-Domain QA Dataset including question answering data in the fields of coding, mathematics, and natural language, i.e., Text2Code~\cite{liu2023mftcoder}, Orca Math~\cite{mitra2024orca}, and a combination of Alpaca-cleaned~\cite{sialpaca} and Instinwild~\cite{xue2023instruction} datasets. The statistics of this dataset are shown in Table~\ref{tab:multi-domain-qa-dataset}. We select the checkpoint with the lowest validation loss and evaluate the model's performance on the test set using perplexity (PPL) as the metric, with lower perplexity indicating better performance.

\begin{table}[t]
    \centering
    \caption{Performance for the Code Completion task in the CRT dataset.}
    \resizebox{\linewidth}{!}{
    \begin{tabular}{lcccccc}
        \toprule
        \pmb{Method} & \pmb{Python} & \pmb{Java} & \pmb{C++} & \pmb{JS} & \pmb{Go} & \pmb{Avg} \\
        \midrule
        STL & 46.3 & 27.4 & 20.1 & 29.3 & 19.5 & 28.5 \\
        Uniform & 48.2 & 35.4 & 25.6 &  31.1 & 22.0 & 32.4 \\
        GradNorm & 49.4 & 36.0 & 25.0 & 30.5 & \pmb{24.4} & 33.1 \\
        LBTW & 48.2 & 35.4 & 28.7 & 31.7 & 20.7 & 32.9 \\
        FAMO & 47.6 & 35.4 & 23.2 & 32.3 & 22.0 & 32.1 \\
        CoBa & \pmb{50.0} & \pmb{39.0} & \pmb{29.3} & \pmb{32.9} & \pmb{24.4} & \pmb{35.1} \\
        \bottomrule
    \end{tabular}}
    \label{tab:5tasks-codecompletion}
\end{table}

\begin{table}[t]
    \centering
    \caption{Performance for the Unit Test Generation task in the CRT dataset.}
    \resizebox{0.85\linewidth}{!}{
    \begin{tabular}{lcccc}
        \toprule
        \pmb{Method} & \pmb{Python} & \pmb{Java} & \pmb{JavaScript} & \pmb{Avg} \\
        \midrule
        STL & 11.5 & 11.6 & 4.3 & 9.1 \\
        Uniform & 15.3 & 17.7 & 14.6 & 15.9 \\
        GradNorm & 14.0 & 15.2 & 16.5 & 15.2 \\
        LBTW & 11.5 & \pmb{20.1} & 15.2 & 15.6 \\
        FAMO & 10.9 & 11.6 & \pmb{19.5} & 14.0 \\
        CoBa & \pmb{19.1} & 17.7 & 16.5 &   \pmb{17.7} \\
        \bottomrule
    \end{tabular}}
    \label{tab:5tasks-unittest}
\end{table}

\begin{table*}[t]
    \centering
    \caption{Performance for the Code Translation task in the CRT dataset.}
    \resizebox{0.7\linewidth}{!}{
    \begin{tabular}{lccccccc}
        \toprule
        \pmb{Method} & \pmb{Py2Java} & \pmb{Py2C++} & \pmb{Java2Py} & \pmb{Java2C++} & \pmb{C++2Py} & \pmb{C++2Java} & \pmb{Avg} \\
        \midrule
        STL & 44.5 & 57.8 & 51.8 & \pmb{42.7} & 53.7 & 45.7 & 49.4 \\
        Uniform & 62.5 & 56.4 & 59.2 & 28.7 & 60.4 & 58.5 & 54.3 \\
        GradNorm & 63.9 & 55.4 & \pmb{64.6} & 23.2 & 63.4 & 56.7 & 54.5 \\
        LBTW & 66.5 & 52.6 & 61.6 & 28.1 & \pmb{64.63} & 60.4 & 55.6 \\
        FAMO & 63.7 & 56.4 & 62.2 & 25.0 & 61.6 & \pmb{62.8} & 55.3 \\
        CoBa & \pmb{67.3} & \pmb{61.2} & 61.0 & 28.1 & 59.8 & \pmb{62.8} & \pmb{56.7} \\
        \bottomrule
    \end{tabular}}
    \label{tab:5tasks-codetrans}
\end{table*}

\begin{table}[t]\small
    \centering
    \setlength{\tabcolsep}{26pt}
    \caption{Performance for the Code Comment task in the CRT dataset.}
    \begin{tabular}{lc}
        \toprule
        \pmb{Method} & \pmb{BLEU} \\
        \midrule
        STL & 34.6 \\
        Uniform & 34.5 \\
        GradNorm & 34.0 \\
        LBTW & 34.8 \\
        FAMO & 33.8 \\
        CoBa & \pmb{35.4} \\
        \bottomrule
    \end{tabular}
    \label{tab:5tasks-codecomment}
\end{table}

\begin{table}[t]\small
    \centering
    \setlength{\tabcolsep}{26pt}
    \caption{Performance for the Text2Code task in the CRT dataset.}
    \begin{tabular}{lc}
        \toprule
        \pmb{Method} & \pmb{MBPP (Pass@1)} \\
        \midrule
        STL & 41.0 \\
        Uniform & 41.4 \\
        GradNorm & 41.0 \\
        LBTW & \pmb{41.6} \\
        FAMO & 40.2 \\
        CoBa & 41.0 \\
        \bottomrule
    \end{tabular}
    \label{tab:5tasks-text2code}
\end{table}

\section{Experiment Setup}
\label{app:setup}

In this section, we elaborate on the experimental setups for benchmark methods used in our paper.

Regarding the CC and CRT Dataset, our chosen base model is Phi-1.5-1.3B~\cite{li2023textbooks} due to its strong coding power. We fine-tune this model using a cluster of 16 A100 GPUs, with specific parameters set as follows: a learning rate of 5e-6, and a total batch size of 160. For the Code Completion dataset, we ensure uniform sample length by adding padding tokens. In the case of the Code-Related Tasks Dataset, we employ a data pack mode to accommodate its extensive sample size. This technique packs samples and ensures their cumulated length does not exceed the sequence length of the base model, thereby boosting training efficiency~\cite{touvron2023llama,liu2023mftcoder}. In addition, to compare the baseline methods' performance with a larger model, we utilize CodeLlama-13B-Python~\cite{roziere2023code} as the base model in the Code Completion Dataset with a learning rate of 1e-6 and a batch size of 128.

For the XTREME-UP dataset, we select Qwen-1.8B~\cite{bai2023qwen} as our base model since it is a multilingual model. Fine-tuning proceeds on 8 A100 GPUs, with a learning rate of 5e-7, a total batch size of 128, and the adoption of the padding mode. It's crucial to highlight that when replicating FAMO, we utilized a larger learning rate of 5e-6, as the re-normalization and regularization techniques employed by FAMO make the training converge too slowly when the learning rate is 5e-7.

For the Multi-Domain QA dataset, we choose Phi-1.5-1.3B~\cite{li2023textbooks} again because there is code-related data in this dataset. We fine-tune this model using 8 A100 GPUs, with a learning rate of 1e-5, a total batch size of 80, and the adoption of the pack mode.

In regards to hyperparameters used in all methods, the following settings are applied for CoBa: $au$ is set to 5, $N$ is set to $2M$, and $W$ is set to $M$, with $M$ representing the batch number of the validation set. For GradNorm, we assign the asymmetry hyperparameter $\alpha$ a value of 1.5, as this provides the best performance in their respective studies, and utilize the `lm\_head' layer as $\theta_{s}$. In the case of LBTW, we adjust the hyperparameter $b$ to 0.5, again following the best performance guidelines from their research. With FAMO, the settings include a learning rate of 0.025 for the optimizer for the weights $\alpha$, and a weight decay $\gamma$ of 0.01.

Finally, to ensure a fair comparison, we include the early stopping method in our fine-tuning procedure, based on the validation loss ratio averaged over all tasks. The checkpoint with the lowest validation loss ratio is selected for downstream evaluation. 

\section{Results on CRT Dataset}
\label{app:results_code_related}

\begin{figure*}[t]
    \centering
    \includegraphics[width=1\textwidth]{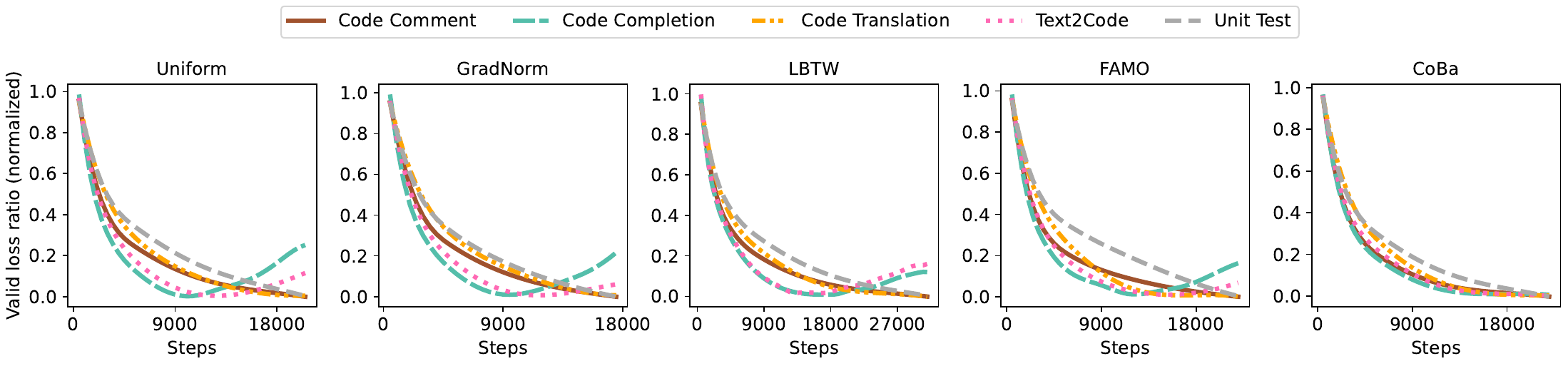}
    \caption{Normalized valid loss ratio of 5 programming tasks on CRT dataset. The x-axis endpoint in each figure marks the early stopping point. For better visualization, we apply Min-Max Normalization to the validation loss ratios for each task, which involves subtracting the minimum value and then dividing by the range between the maximum and minimum values.}
    \label{fig:5tasks-loss}
\end{figure*}

We further investigate the performance of all methods on the Code-Related Tasks (CRT) Dataset. Here we split the tasks based on the specific coding requirements, rather than the programming language. Note that the sample size of this dataset is much larger than the other two, and the high complexity of GradNorm$^*$ and FAMO$^*$ precludes their use on this dataset. The results, distributed across Tables~\ref{tab:5tasks-codecompletion} to~\ref{tab:5tasks-codecomment}, indicate that CoBa surpasses other SOTA methods in almost all tasks, excluding Text2Code. In particular, CoBa stands out in the Code Completion and Unit Test Generation tasks, recording substantial relative average Pass@1 enhancements of at least 6\% and 13\%, respectively. Furthermore, as depicted in Figure~\ref{fig:5tasks-loss}, CoBa not only avoids early divergence in Code Completion and Text2Code tasks but also expedites convergence in the remaining tasks, affirming its efficacy in achieving convergence balance and boosting MTL capabilities. Conversely, other methods aimed at balancing convergence, such as GradNorm, LBTW, and FAMO, exhibit erratically across different tasks, often failing to prevent overfitting in Code Completion and Text2Code tasks. Their performance is sometimes even inferior to STL, which learns each task separately, highlighting a potential limitation of these methods compared to the robustness of CoBa.


\begin{table}[t]
    \centering
    \caption{Performance in the Multi-Domain QA dataset.}
    \resizebox{0.85\linewidth}{!}{
    \begin{tabular}{lcccc}
        \toprule
        \pmb{Method} & \pmb{Code} & \pmb{Math} & \pmb{NL} & \pmb{Avg} \\
        \midrule
        Uniform & 2.0103 & 1.3629 & 3.9749 & 2.2167 \\
        GradNorm & 2.0103 & 1.3623 & 3.9749 & 2.2162 \\
        GradNorm$^{*}$ & 2.0103 & 1.3623 & 3.9749 & 2.2162 \\
        LBTW & 2.0109 & 1.3652 & 3.8768 & 2.1997 \\
        LBTW$^{*}$ & 2.0103 & 1.3619 & 3.8690 & 2.1961 \\
        FAMO & 2.0119 & 1.3633 & 3.9236 & 2.2078 \\
        FAMO$^{*}$ & 2.0113 & 1.3597 & 3.9157 & 2.2043 \\
        CoBa & \pmb{2.0075} & \pmb{1.3568} & \pmb{3.8574} & \pmb{2.1902} \\
        \bottomrule
    \end{tabular}}
    \label{tab:multi-domain-qa-results}
\end{table}

\begin{figure*}[htb]
    \centering
    \subfigure{
        \includegraphics[width=0.6\textwidth]{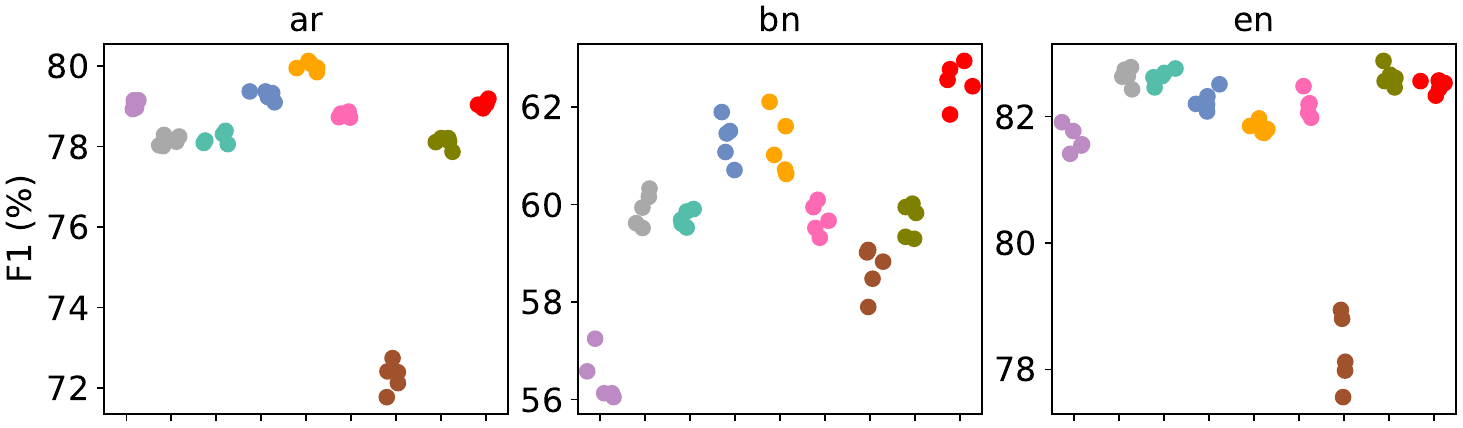}
    }
    \vspace{-4mm}
    \\
    \subfigure{
        \includegraphics[width=0.8\textwidth]{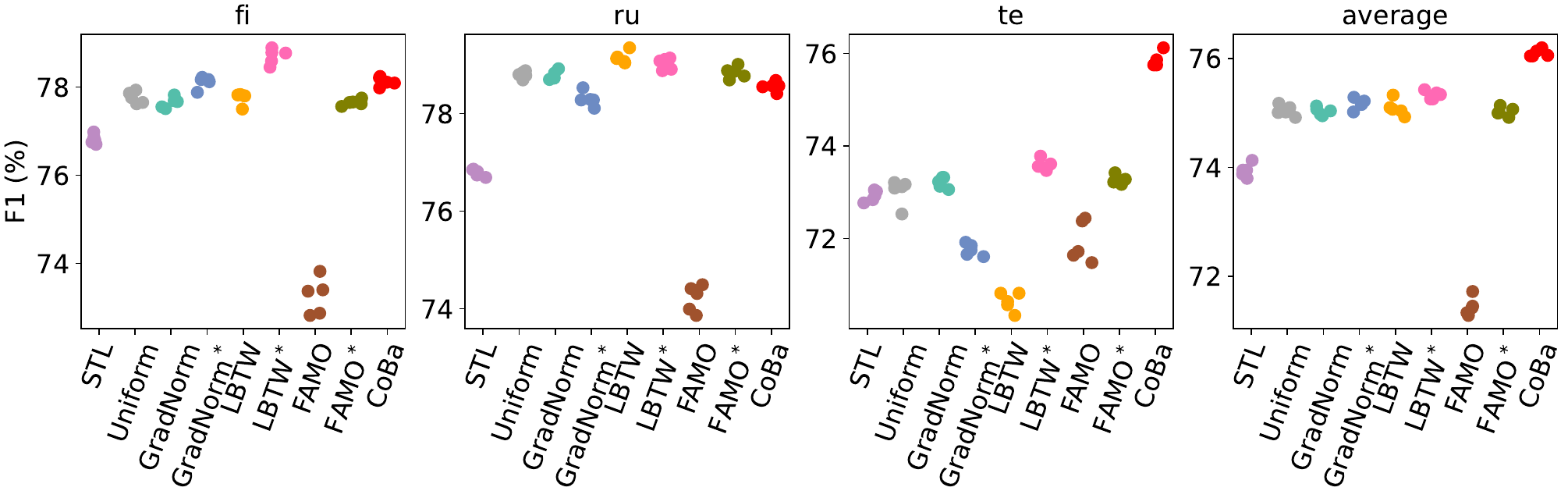}
    }
    \caption{Experimental results on XTREME-UP dataset with 6-tasks setting.}
    \label{fig:xtreme-up-6tasks}
\end{figure*}

\begin{figure*}[t]
    \centering
    \includegraphics[width=1.0\textwidth]{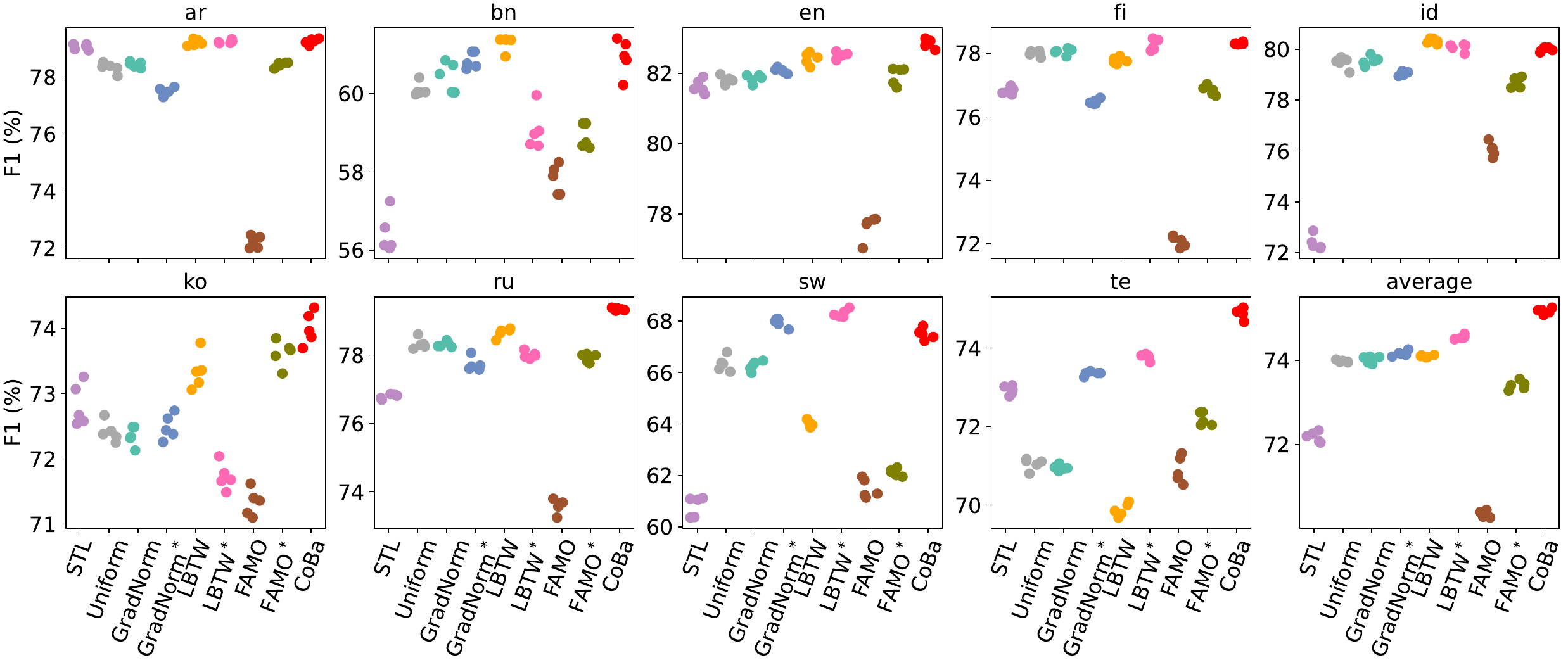}
    \caption{Experimental results on XTREME-UP dataset with 9-tasks setting.}
    \label{fig:xtreme-up-9tasks}
\end{figure*}

\section{Results on Multi-Domain QA Dataset}
\label{app:results_multi_domain_qa}
The results are summarized in Table~\ref{tab:multi-domain-qa-results}, demonstrating that CoBa consistently achieves the lowest PPL across all three tasks, underscoring its robustness in handling datasets with high diversity. Compared to the second-best baseline (i.e., LBTW$^{*}$), CoBa reduces the average perplexity by $0.0059$. Moreover, when compared to the worst-performing baseline (i.e., Uniform), CoBa shows an average perplexity reduction of $0.0265$. The experimental results demonstrate the effectiveness of CoBa on multi-task datasets in different domains. In this experiment, we set a larger learning rate compared to previous experiments. Our findings reveal that the performance of FAMO is comparable to FAMO*. This suggests that FAMO is sensitive to the learning rate.

\end{document}